\documentclass[journal]{IEEEtran}

\usepackage{epsfig}
\usepackage{graphicx}
\usepackage{amsmath}
\usepackage{amssymb}

\usepackage{color}
\usepackage{soul}
\usepackage[utf8]{inputenc}
\usepackage[pagebackref=true,breaklinks=true,letterpaper=true,colorlinks,bookmarks=false]{hyperref}

\usepackage[figuresright]{rotating}
\usepackage{textcomp,subfigure,algorithm,algorithmic,multirow,upgreek,tabularx}
\usepackage{graphics,gensymb}
\usepackage{bm,cases,threeparttable}

\usepackage{setspace}

\graphicspath{{Figures/}}
\usepackage{booktabs}

\usepackage{amsfonts}

\usepackage{caption}
\usepackage{enumitem}
\setitemize{noitemsep,topsep=0pt,parsep=0pt,partopsep=0pt}

\usepackage{textcomp}

\usepackage{makecell}

\usepackage{color}

\ifCLASSINFOpdf
\else
\fi

\begin{document}
% \title{PanoGAN: Feedback Guided GANs for Cross-View Panorama Image Synthesis}
\title{Cross-View Panorama Image Synthesis}

\author{
        Songsong~Wu,
        Hao~Tang,
        Xiao-Yuan~Jing,
        Haifeng~Zhao,
        Jianjun Qian,
        Nicu~Sebe,~\IEEEmembership{Senior Member,~IEEE,}
        and~Yan~Yan,~\IEEEmembership{Senior Member,~IEEE}% <-this % stops a space

\thanks{Songsong Wu is with the School of Computer Science, Guangdong University of Petrochemical Technology, Maoming 525000, China. (e-mail:sswuai@126.com)}
\thanks{Hao Tang is with the Department of Information Technology and Electrical Engineering, ETH Zurich, Zurich 8092, Switzerland (e-mail: hao.tang@vision.ee.ethz.ch)}
\thanks{Xiao-Yuan Jing is with the School of Computer Science, Guangdong University of Petrochemical Technology, Maoming 525000, China, the School of Computer Science, Wuhan University, Wuhan 430072, China, and State Key Laboratory for Novel Software Technology, Nanjing University, Nanjing 210023, China. (e-mail: jingxy\_2000@126.com)}
\thanks{Haifeng Zhao is with the School of Software Engineering, Jinling Institute of Technology, 211169, Nanjing, China, and also with Jiangsu Hoperun Software Co. Ltd., Nanjing, 210012, China. (e-mail: zhf@jit.edu.cn)}
\thanks{Jianjun Qian is with the School of Computer Science and Engineering, Nanjing University of Science and Technology, Nanjing 210094, China. (e-mail: csjqian@njust.edu.cn)}
\thanks{Nicu Sebe is with the Department of Information Engineering and Computer Science (DISI), University of Trento, Trento 38123, Italy. (e-mail: sebe@disi.unitn.it)}
\thanks{Yan Yan is with the Department of Computer Science, Illinois Institute of Technology, Chicago 60616, USA. (e-mail: yyan34@iit.edu)}
\thanks{Songsong Wu and Hao Tang contributed equally to this work. Corresponding author: Xiao-Yuan Jing}
}

%\markboth{IEEE Transactions on Image Processing}%
\markboth{IEEE Transactions on Multimedia}%
{Shell \MakeLowercase{\textit{et al.}}: Bare Demo of IEEEtran.cls for IEEE Transactions on Magnetics Journals}

% for Transactions on Magnetics papers, we must declare the abstract and
% index terms PRIOR to the title within the \IEEEtitleabstractindextext
% IEEEtran command as these need to go into the title area created by
\IEEEtitleabstractindextext{%

%------------------- Abstract ----------------------------------------------------------------------------------
\begin{abstract}
In this paper, we tackle the problem of synthesizing a ground-view panorama image conditioned on a top-view aerial image, which is a challenging problem due to the large gap between the two image domains with different view-points. Instead of learning cross-view mapping in a feedforward pass, we propose a novel adversarial feedback GAN framework named PanoGAN with two key components: an adversarial feedback module and a dual branch discrimination strategy.
First, the aerial image is fed into the generator to produce a target panorama image and its associated segmentation map in favor of model training with layout semantics.
Second, the feature responses of the discriminator encoded by our adversarial feedback module are fed back to the generator to refine the intermediate representations, so that the generation performance is continually improved through an iterative generation process.
Third, to pursue high-fidelity and semantic consistency of the generated panorama image, we propose a pixel-segmentation alignment mechanism under the dual branch discrimiantion strategy to facilitate cooperation between the generator and the discriminator.
Extensive experimental results on two challenging cross-view image datasets show that PanoGAN enables high-quality panorama image generation with more convincing details than state-of-the-art approaches. The source code and trained models are available at
\url{https://github.com/sswuai/PanoGAN}.
\end{abstract}

}

\maketitle

\IEEEdisplaynontitleabstractindextext

% \fi
\IEEEpeerreviewmaketitle

%------------------- Section 1: Introduction ----------------------------------------------------------------------------------
\section{Introduction}
Cross-view image synthesis aims to translate images from a given view to a novel view by building a semantic-preserving mapping across distinct views. Limited overlap in the field of views causes significantly different appearances of cross-view images, making the task very challenging. As several pioneer works extend image translation approaches to this new task \cite{Zhai_cvpr17,Yi_Zhou_bmvc17},  improvements in model capability and deeper understandings of Generative Adversarial Networks (GANs) are achieved for cross-view image synthesis, which benefits other vision tasks such as 3D object translation \cite{yang2015weakly}, cross-view image matching \cite{Sixing_Hu_cvpr18,regmi2019bridging}, and geo-localization \cite{Scott_Workman_iccv15,tian2017cross}. Thus, cross-view image synthesis has been a core frontier of academic research in computer vision.

A typical cross-view image generation task is to translate scene images of a certain geographical location between the top view and the ground view. Despite recent methods \cite{Krishna_cvpr18,regmi2019bridging,tang_cvpr19,tang2019local} have achieved promising progress to address the problem, they only consider a single camera-angle ground-view image rather than $360$-degree panorama image. In this paper, we address the task of generating a ground-view panorama image by given a top-view aerial image, which is considerably difficult due to the large domain gap between the two views and
+notable difference in image resolution between aerial and panorama images. By following the setting of previous work \cite{Krishna_cvpr18}, we assume that ground-truth panorama segmentation maps are available in the training phase to provide semantic layout information, while unavailable in the inference phase.

Existing cross-view image translation methods struggle to synthesize realistic and semantically meaningful panorama images due to two limitations:
(i) Existing methods generate target images through a feedforward network architecture that actually is a one-off process. Consequently, the generator is unable to refine the synthesized images despite performance evaluation results are provided by the discriminator. It is well known that human beings are good at achieving their purposes through continuous correction of their behavior guided by performance feedback. Inspired by this insight, one may expect to synthesize panorama images in an iterative process.
(ii) Previous GANs based models only use the intermediate feature from the discriminator to distinguish real and fake target images. This discriminator’s response could only implicitly affect the generation process through optimizing the generator. It is desirable to provide the generator with correction guidance information from the discriminator's intermediate feature to facilitate the cooperation between generator and discriminator.

%---------------------------Fig: pipeline of PanoGAN ------------
\begin{figure*}[!t]
    \centering
    \includegraphics[width=1\linewidth]{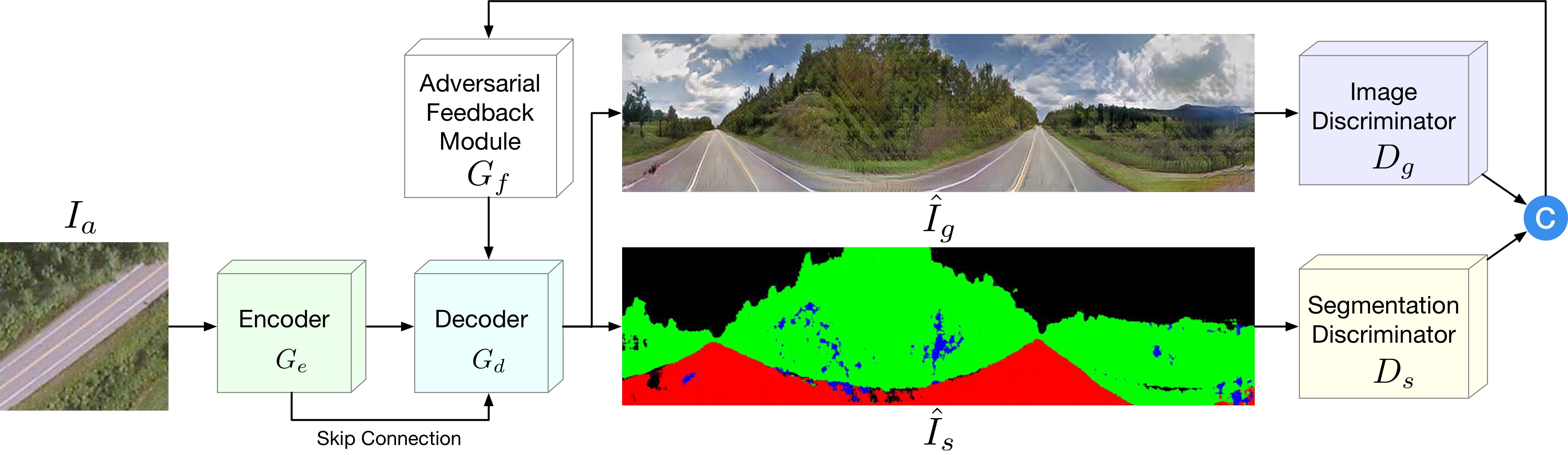}
	\caption{Overview of the proposed PanoGAN network architecture consisting of an encoder $G_e$, a decoder $G_d$, an image discriminator $D_g$, a segmentation discriminator $D_s$, and an adversarial feedback module $G_f$. All these components are trained in an end-to-end way so that each component can benefit from the others. Note that the decoder $G_d$ produces a 6-channel output of which the first 3-channels are treated as the generated panorama image $\hat{I}_g$ and the remaining 3-channels are treated as the generated segmentation map $\hat{I}_s$. The symbol $\textcircled{c}$ denotes channel-wise concatenation.}
	\label{fig:architecture}
\end{figure*}
%---------------------------Fig End: pipeline of PanoGAN ------------

Based on the above analysis, we present in this paper a novel conditional GANs with adversarial feedback learning, termed as PanoGAN, to address the cross-view panorama synthesis task. As shown in Fig.~\ref{fig:architecture}, our model is an iterative generation system consisting of a feedforward generation step and a feedback correction step. The idea is inspired by the well-known truth that human beings are good at continuously correcting their behaviors during filling a task to approach the goal gradually. Specifically, the model takes the input of an aerial image $I_a$ and outputs a panorama image $\hat{I}_g$ and a panorama segmentation $\hat{I}_s$ simultaneously. The intermediate discrimination features of current outputs are transformed through our proposed adversarial feedback module (AFM) and then fed into the generator for the next generation.
In this way, the target panorama image is continuously refined through an iterative generation process guided by the correction information provided by the feedback discriminators' response.

We conduct extensive experiments on two challenging datasets, i.e., CVUSA \cite{workman2015wide} and OP \cite{regmi2019bridging}.
The experiments show that PanoGAN achieves significantly better generation performance than exiting competitive models for cross-view image generation.

In summary, the contributions of our paper are:
\begin{itemize}[leftmargin=*]
  \item  We propose a novel conditional GAN model termed PanoGAN that generates 360-degree ground-view panorama images from top-view aerial images by emphasizing the cooperation and competition of generator and discriminator. We also introduce an adversarial feedback module in PanoGAN to pass the discrimination information from discriminator to generator. As a result, continual improvements in image quality could be achieved through the iterative image generation process, which exploits discrimination feedback as  powerful correcting supervision.
  \item We develop a new discrimination mechanism to pursue high-fidelity and semantic consistency in the generated panorama images. With the extracted multi-scale pyramid features of the panorama images and segmentation maps, our proposed pixel-semantic alignment loss helps the discriminator accurately explore the layout semantics of the target images. Moreover, the discrimination features of panorama images and segmentation maps provide valuable supervision as feedback to the generator for continuously improving synthesis performance.
  \item Extensive experimental results verify the effectiveness of the proposed method both in terms of subjective visual realness and objective qualitative scores. We demonstrate and establish new state-of-the-art performances on two challenging benchmarks, CVUSA~\cite{workman2015wide} and OP~\cite{regmi2019bridging}.
\end{itemize}

%%%--------------------------------Fig: Model overview -------------------------------------------
\begin{figure*}[th]
    \centering
    \includegraphics[width=1\linewidth]{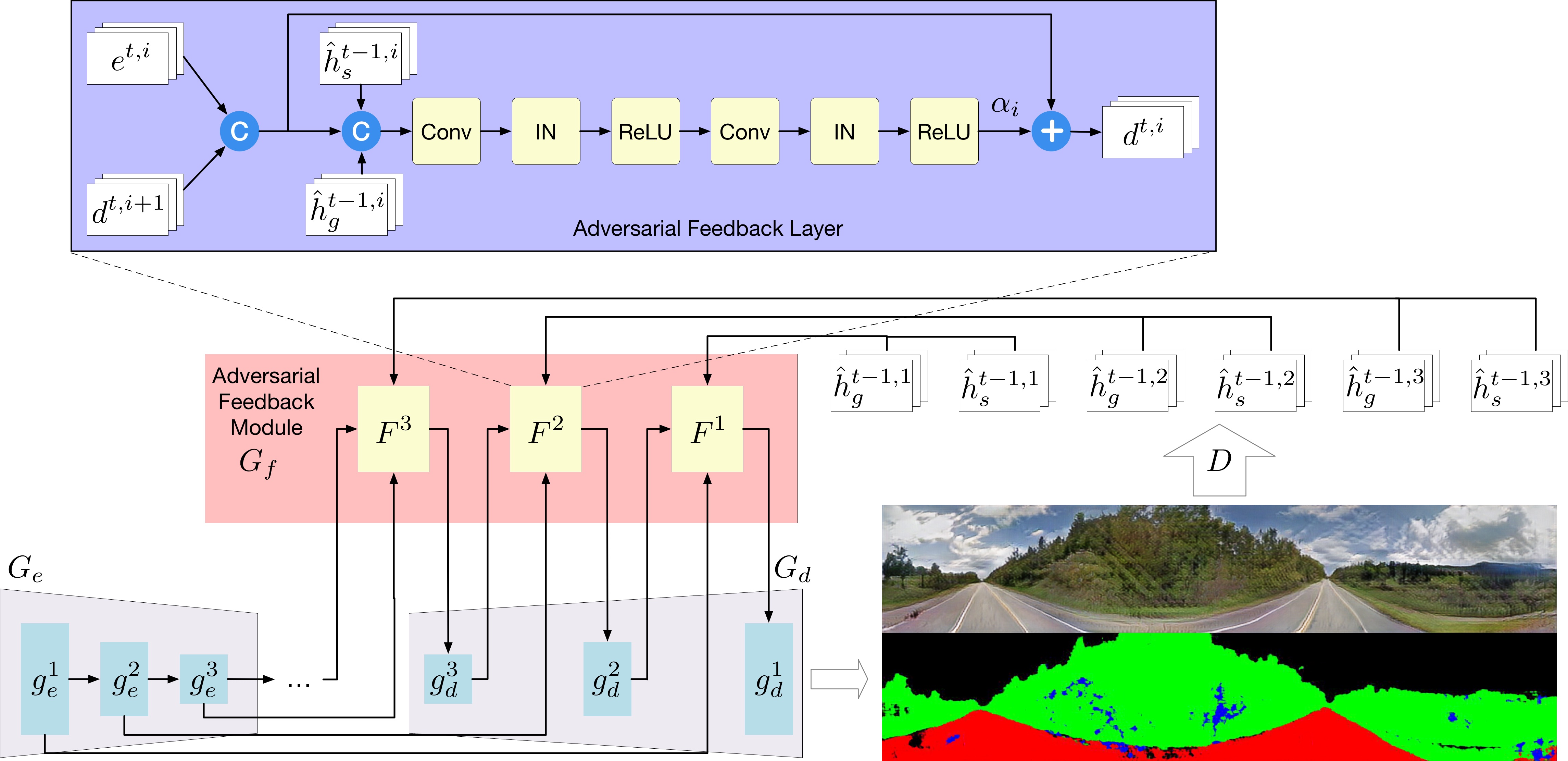}
    \caption{Overview of our iterative generation using the proposed Adversarial Feedback Module (AFM). In the $t$th iteration,the  encoder feature $e^{t,i}$ produced by the encoder layer $g_e^i$, the decoder feature $d^{t,i+1}$ produced by the decoder layer $g_d^{i+1}$, the discriminator features $\hat{h}_g^{t-1,i}$ and $\hat{h}_s^{t-1,i}$ extracted from the panorama image and the segmentation map produced in the previous iteration are fused by the proposed adversarial feedback layer $F^i$ into a novel feature $d^{t,i}$, which is then treated as the input feature of the decoder layer $g_d^i$.
    Each adversarial feedback layer consists of stacked convolution-normalization-ReLU blocks as shown in the upper half part of the figure, in which $\alpha_i$ is the weighting parameter for feature fusion. The symbols $\textcircled{c}$ and $\textcircled{+}$ denote channel-wise concatenation, and element-wise addition, respectively.}
    \label{fig:layer}
\end{figure*}
%%%--------------------------------Fig End: Model overview -------------------------------------------

%------------------- Section 2: Related Work ----------------------------------------------------------------------------------
\section{Related Work}
\label{sec_rel}
\noindent\textbf{Generative Adversarial Networks (GANs)} have two important components, i.e., a generator and a discriminator, which are optimized in an adversarial way to achieve a balance.
Recently, GANs have shown the capability of generating realistic images in different tasks such as human face~\cite{WWang_tmm2020}, human pose~\cite{guo2019fusegan}, animal and plant~\cite{zhang_tpami2019_stackgan++} and scene images~\cite{zhou2019branchgan}. In order to generate user-specific images, a conditional GAN (cGAN)~\cite{mirza2014conditional} usually injects into vanilla GAN model external information such as human pose \cite{tang2020xinggan,gu2020toward}, class labels \cite{wang2020learning,wu2019relgan}, and text descriptions \cite{zhang2017stackgan}.
Pix2pix~\cite{isola2017image} provides a general image-to-image translation framework using cGAN, which is demonstrated to be effective for tasks such as label-to-scene, day-to-night, and edge-to-photo. Thus, we adopt Pix2pix as the backbone of our architecture in this work.

\noindent\textbf{Cross-View Image Synthesis}
aims to translate images from a given view to a novel view. The early works focus on cross-view generation for object images \cite{Dosovitskiy_tpami17,li2018hole} and scene images \cite{Zhou_eccv16,liu2020exocentric} with large overlap of view. Recently, research attention focuses on scene image synthesis between top-view and ground-view, in which there is no overlap between the views. Krishna et al. \cite{Krishna_cvpr18} propose to translate images between top-view and ground-view via multi-task learning that generates images and their corresponding segmentation maps simultaneously. In~\cite{regmi2019bridging}, the problem of image matching between different views is addressed by generating aerial images from panorama images. Tang et al. \cite{tang_cvpr19} propose a multi-channel selection mechanism to address the issue of multi-scale objects in view transformation. Different from synthesizing one part of ground-view panorama image as in ~\cite{Regmi_cvpr18} and \cite{tang_cvpr19}, the whole panorama image is considered firstly in~\cite{Zhai_cvpr17} in which a ground-level panorama image is generated with its label map as supervision.  Krishna et al. \cite{regmi2019bridging} address the problem of image matching between different views by generating aerial images from panorama images.
Our model differs from previous approaches in that it synthesizes the target image iteratively with the guidance of the feedback features from the discriminator.

\noindent\textbf{Adversarial Feedback Learning} aims to increase the performance of GANs by guiding the network with the evaluation information of current performance. Many tasks such as image classification \cite{Sanath_Narayan_eccv20}, human pose estimation \cite{Joao_Carreira_cvpr16}, and image generation \cite{Shama_iccv19,Huh_cvpr19} have benefited from the feedback learning mechanism.
For instance, Huh et al. \cite{Huh_cvpr19} propose to feedback discriminator's spatial output response to the generator in a GAN framework, being as such the most related work to us. The feedback is achieved by modifying the bottom-neck generator feature with the score map of discriminators in a conditional batch normalization manner.
Shama et al. \cite{Shama_iccv19} is another related work that extends the feedback learning from the training process to the inference process.
Compared with both methods,
our method sends multiple discriminator features back to each decoder layer for better using the feedback information.
Moreover, our method uses a novel multi-scale discriminator to aid the generator and makes the feedback learning in an end-to-end manner.

\noindent\textbf{Multi-Scale Discriminator.} Conventional GANs use the top-layer feature of the discriminator to make a judgment of real or fake based on single-scale local patches. Recently, the features provided by different layers were exploited to enhance the ability of the discriminator in the applications of semantic image generation \cite{Wang_cvpr18}, semantic segmentation \cite{Lin_cvpr17,Long_cvpr15}, object detection \cite{Kim_icml17}, and cross-view image matching \cite{TLin_cvpr17}.
There are two ways to compute the multi-scale discriminator feature:
apply discriminators with different scales to extract multi-scale features (e.g., multi-scale PatchGAN) or
use a feature pyramid discriminator to extract multi-scale features from different layers (e.g., \cite{TLin_cvpr17}).
Inspired by \cite{Liu_nips19}, we design a novel dual branch discriminator with multi-scale pyramids to empower the discriminator under a pixel-semantic alignment mechanism.  Besides, the multi-scale discrimination features are also used as feedback guidance information to help the generator refine the resulted images.

%------------------- Section 3: Formulation ----------------------------------------------------------------------------------
\section{The Proposed PanoGAN}
\label{formulation}
%%%--------------------------------Fig: Module AFM -------------------------------------------
\begin{figure*}[htp]
    \centering
    \includegraphics[width=1\linewidth]{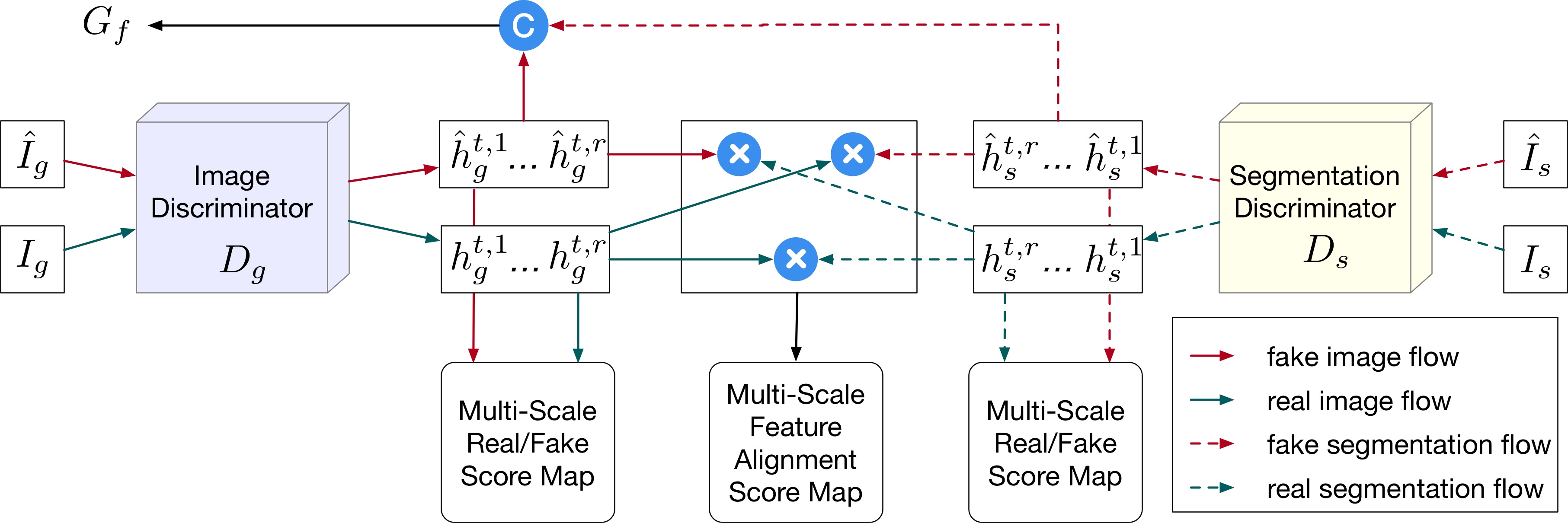}
    \caption{Framework of the proposed pixel-semantic alignment mechanism. Image discriminator features and segmentation discriminator features are used for real/fake determining and pixel-semantic alignment. They are also concatenated as the feedback information to the generator. The symbols $\textcircled{c}$ and $\otimes$ denote channel-wise concatenation, and element-wise multiplication, respectively.}
    \label{fig:G_D}
\end{figure*}
%%%--------------------------------Fig End: MOdul AFM -------------------------------------------

\subsection{Problem Formulation}
\label{sub:method-problem}
We focus on learning a translation mapping from the top-view domain to the ground-view domain, to synthesize a panorama image $I_g {\in} \mathbb{R}^{m \times n}$ conditioned on an input aerial image $I_a {\in} \mathbb{R}^{p \times q}$.
The translation mapping is supposed to satisfy two demands for the generated images: semantic consistency and image fidelity.
For semantic consistency, the generated panorama images should preserve semantic information of the layout and objects in the input aerial images because both aerial and panorama images depict the same location.
For image fidelity, the generated panorama image should maintain most of the local details and meanwhile introduce as few artifacts as possible.

The task is challenging due to no overlap in the field of view between the two domains. Moreover, only the aerial images are available for panorama image generation, making the task harder.
As the semantic layout information in the segmentation map can be used to guide the training of cross-view transformation system \cite{Krishna_cvpr18}, we also provide the segmentation maps of panorama images in the training set. Note that our task is different from \cite{tang_cvpr19} in that segmentation map is only used for model training, while \cite{tang_cvpr19} use segmentation map for both the model training and model inference.

\subsection{Overall Network Architecture}
Our model is an iterative process consisting of a feedforward generation step and a feedback correction step. This architecture helps to promote generator-discriminator cooperation in the generation process and continuously improves the image quality of the target image. The overall framework of our panorama image generation system is shown in Fig.~\ref{fig:architecture}. Denote by $I_a$ the aerial image that is fed into the generator to extract semantic features associated with top-view. The generator outputs a pair of ground-view panorama images $\hat{I}_g$ and segmentation map $\hat{I}_s$. In the feedback correction step, the image discriminator $D_g$ and the segmentation discriminator $D_s$ feed their features into the proposed adversarial feedback module $G_f$ to learn the feedback feature. In the feedforward generation step, the encoding feature of the encoder $G_d$ and feedback feature are integrated and fed into the decoder $G_d$ to generate a refined version of the panorama image and its segmentation map.

Our adversarial feedback module bridges the generator and discriminator as the discrimination feature is directly involved in the image generation process.  As a result, the generator and discriminator not only compete against each other but also are mutually reinforcing, enabling the generation performance to be continually improved.

\subsection{Generator Backbone}
As shown in Fig.~\ref{fig:layer}, the generator consists of an encoder $G_e$ with multiple encoding layers and a decoder $G_d$ with the same number of decoding layers. Each encoding layer is a sequential operation of convolution-normalization-ReLU and each decoding layer is a sequential operation of deconvolution-normalization-ReLU. There is an encoding-feedback-decoding bridge among the corresponding encoding layer, adversarial feedback (AF) module, and decoding layer, which is analogous to U-Net network architecture. The 6-channel output of $G_d$ is considered as the channel-wise concatenation of the panorama image (the first 3 channels) and its corresponding segmentation map (the last 3 channels).

In the training phase, the generated segmentation map takes part in adversarial learning accompanied by the ground-truth segmentation map, which provides supervision information for cross-view transformation. In the inference phase,  the generated segmentation map provides feedback features with semantic information at the object-class.

Note that our generator differs slightly from X-Fork~\cite{Krishna_cvpr18} or X-Seq~\cite{Krishna_cvpr18} in that panorama image and its segmentation map are produced simultaneously by a single decoder rather than two parallel network branches in X-Fork or two sequential sub-networks in X-Seq.

\subsection{Iterative Generation by Adversarial Feedback Learning}
\label{subsect: generator}

Besides the encoder $G_e$ and decoder $G_d$, we propose an Adversarial Feedback Module (AFM) $G_f$ to achieve iterative image generation with correction guidance fed back from the discriminators. As shown in Fig.~\ref{fig:layer}, $G_f$ is comprised of multiple layers $F$s, each of which contains two stacked Convolution-Normalization-ReLU blocks. There is a skip connection between the corresponding encoding layer and the AFM layer to fuse the semantic encoding information and feedback correction information.

Specifically, we assume that the current iteration order is $t$ and focus on the input to be fed into the $i$-th decoding layer $g_d^i$. Let $e^{t,i}$ be the encoding feature produced by $g_e^i$, $d^{t,i+1}$ be the decoding feature produced by $g_d^{i+1}$, $\hat{h}_g^{t-1,i}$ and $\hat{h}_s^{t-1,i}$ be the feedback features of $\hat{I}_g$ and $\hat{I}_s$ obtained in the $(t{-}1)$-th iteration, respectively.
Firstly, the encoding feature and decoding feature are concatenated along the channel axis as $e_i \textcircled{c} d^{t-1}_i$, which is then concatenated with feedback features along the channel axis to be the input of AFM layer $F^i$ for feature transformation. By introducing skip connection between the two ends of AFM layer $F^i$, fused feature $d^{t,i}$ is obtained by elementary addition of encoding-decoding feature and feedback correction feature, which is fed into $g_d^i$ for the following feature decoding.
The feature fusion operation in the AFM layer can be formulated as,
\begin{equation}
  d^{t,i} = \underbrace{\alpha_i F_i \big(\left(e^{t,i}\textcircled{c}d^{t,i+1}\right) \textcircled{c} h_{g}^{t-1,i} \textcircled{c} h_{s}^{t-1,i}\big)}_{\text{Adversarial Feedback Learning}} \textcircled{+} \underbrace{\big(e^{t,i}\textcircled{c}d^{t,i+1}\big)}_{\text{Skip Connection}},
  \label{eq:fbl}
\end{equation}
where $\textcircled{c}$ is the concatenation operation along the channel axis, $\textcircled{+}$ is the elementary addition operation of feature tensors, and $\alpha_i$ is a weight parameter that determines the importance of feedback features.

The proposed iterative generation framework benefits cross-view panorama image generation in two aspects: (i) As the generator can be provided by feedback correction guidance during the generation process, it can gradually refine the resulted panorama image through iteration. (ii) The intermediate discrimination feature is involved in panorama generating, promoting the cooperation of the generator and discriminator besides their competition. Therefore, semantics preservation in image content and realism in image appearance can be expected in the synthesized panorama image.

%%%-------------------TABLE: experimental setups -------------------------------------------------=
\begin{table*}[htp] \small
\centering
\caption{Interfaces of all the comparison methods in the experiments. (aerial denotes aerial image, pano-img denotes panorama image, pano-seg denotes panorama segmentation map)}
  \begin{tabular}{p{2.5cm}|p{2cm}<{\centering}|p{3cm}<{\centering}|p{5cm}<{\centering}} \toprule
Method    &  Model Input  &  Model Output &  Images for Model Training     \\
   \hline
   Pix2Pix~\cite{isola2017image}       & \multirow{5}{*}{Aerial Image} & Panorama Image  &  Aerial Image, Panorama Image  \\
   \cline{1-1} \cline{3-4}
   X-Fork~\cite{Krishna_cvpr18}        &  & \multirow{4}{*}{\makecell[c]{Panorama Image,\\ Panorama Segmentation}} &
                 \multirow{4}{*}{\makecell[c]{Aerial Image, \\ Panorama Image, \\ Panorama Segmentation }}  \\
   X-Seq~\cite{Krishna_cvpr18}         &    &    &       \\
   SelectionGAN~\cite{tang_cvpr19}  &    &    &       \\
   PanoGAN (Ours)       &    &    &       \\
   \bottomrule
\end{tabular}
\label{tab:interface}
\end{table*}
%%% -------------------TABLE End: experimental setups----------------------------------------------------

\subsection{Discrimination with Pixel-Semantic Alignment}
\label{subsect: discriminator}
We propose a novel discrimination mechanism to guide the generator to synthesize panorama images with accurate semantic layout and high-fidelity details. As shown in Fig.~\ref{fig:G_D}, there are two discriminators, i.e., the image discriminator $D_g$ and the segmentation discriminator $D_s$ that share the same structure but with their respective distinct network parameters. Inspired by recent advances in designing the discriminator of the GAN model~\cite{Wang_cvpr18, Liu_nips19} that use multi-scale features for image discrimination, we adopt the feature pyramid discriminator proposed in~\cite{Liu_nips19} as the backbone of our discriminator.
On this basis, we integrate fake image identification, semantic feature alignment, and feedback feature construction in the proposed discrimination mechanism.
Concretely, the impact of the discrimination process on the generator in PanoGAN is three-fold: (i) the two discriminators guide the generator to accurately estimate the data distributions of panorama image and segmentation map through adversarial learning; (ii) intermediate discriminator features of panorama image and segmentation map are forced to mutually match through a pixel-semantic alignment loss; (iii) the multi-scale discriminator features of fake panorama image and segmentation map are fed back to the generator layer-by-layer to correct previous generations for further improving the generation results.

\noindent\textbf{Multi-Scale Feature Representation}. Inspired by the feature pyramid discriminator proposed in~\cite{Liu_nips19}, our image discriminator $D_g$ and segmentation discriminator $D_s$ distinguish the real or fake images and segmentation maps based on multi-scale features, respectively.
The advantage is that high-level semantic information and low-level appearance property can be captured without a heavy computational burden.
In particular, the discriminator layers transfer their focus from low-level appearance to high-level semantics.
The bottleneck feature maps are added sequentially with the previous feature maps to provide a multiple-scale description of images. Moreover, up-sampling and one-layer convolution are used to adapt the size of the bottle-neck feature in the combination process.

The multi-scale discrimination features affect the network in three aspects: (i) they are used to compute the real/fake score map for adversarial competition between the generator and the discriminator during the model training stage; (ii) they provide feature embeddings for semantic alignment; (iii) the discriminator features of fake image and segmentation map are concatenated in channel axis to be fed back to the generator for the following generation loop.

\noindent\textbf{Pixel-Semantic Alignment.} We propose a novel pixel-semantic alignment mechanism to facilitate appearance-semantic matching through interaction between panorama images and their segmentation maps. The main idea is to use layout semantics in segmentation maps as supplementary information for accurate distinguishment between fake or real panorama images, meanwhile use the appearance information in panorama images as supplementary information to distinguish fake segmentation maps from the real ones.

Assume that there are $r$ discriminator layers, and denote by $h^{t,i}_{g}$ and $h_s^{t,i}$ $(i{=}1,2,\ldots,r)$ the multi-scale feature maps of real image and segmentation map, respectively, where $r$ is the total number of layers in the discriminator.
Similarly, $\hat{h}^{t,i}_g$ and $\hat{h}^{t,i}_s$ represent the multi-scale features of the fake image and segmentation map.
The alignment score map at $i$-th layer can be expressed as

\begin{equation}
\begin{aligned}
  \hat{S}_{g}^{t,i} & = &   \hat{h}_g^{t,i} \otimes h_s^{t,i}, \\
  \hat{S}_{s}^{t,i} & = &  \hat{h}_s^{t,i} \otimes h_g^{t,i}, \\
  S_{g}^{t,i} & = & S_{s}^{t,i} =  h_g^{t,i} \otimes h_s^{t,i},
 \end{aligned}
\label{eq:align_score}
\end{equation}
where $\otimes$ denotes the element-wise multiplication between two feature tensors.

Eq.~\eqref{eq:align_score} shows that the multi-scale feature alignment score map is comprised of three components including the score map $\hat{S}_{g}^{t,i}$ of fake panorama image and real segmentation map, the score map $\hat{S}_{s}^{t,i}$ of real panorama image and fake segmentation map, and the score map $S_{g}^{t,i}$ or $S_{s}^{t,i}$ of real panorama image and real segmentation map. Note that the alignment score map is determined by both the generator and discriminator, thus its gradient plays a role to update the discriminator besides the generator. This differs notably from the multi-scale feature discrimination mechanism proposed in \cite{Liu_nips19} in that only the generator is updated based on the match score of the fake image and the real segmentation map.

The proposed dual branch discrimination strategy enjoys two merits compared with \cite{Liu_nips19}. Firstly, besides the real/fake score map, richer alignment constraints are used to update both the generator and the discriminator. Secondly, our alignment score map is computed by seamlessly integrating it into the discrimination procedure, ultimately benefiting the panorama image generation task.

\subsection{Optimization  Objective}
Assume that our PanoGAN generates a panorama image through $T$ feedback loops, then the final loss of PanoGAN is the average of model losses of all the loops. Three optimization loss functions are considered in our PanoGAN in an end-to-end fashion, i.e., adversarial loss, pixel-semantic alignment loss of discriminator features, and $L1$ distance-based reconstruction loss of panorama images and segmentation maps.

\noindent \textbf{Adversarial Loss.} Under the proposed discrimination mechanism, multi-scale real/fake score maps are used to compute the adversarial loss that indicates how real the generated panorama image and segmentation maps are in appearance. Specifically, the image adversarial loss $L_{g}^t$ and the segmentation adversarial loss $L_{s}^t$ at iteration loop $t$ are computed by adding up all the  losses from $r$ discriminator layers, i.e.,
\begin{equation}
\begin{aligned}
  L_{g}^t & =\sum_{i=1}^{r}{E_{I_a, I_g}\big(\text{log}D_g(h_g^{t,i}) \big) + E_{I_a, \hat{I}^t_g}\big(\text{log}(1 - D_g(\hat{h}_g^{t,i}) ) \big)}, \\
  L_{s}^t & =\sum_{i=1}^{r}{E_{I_a, I_s}\big(\text{log}D_s(h_s^{t,i}) \big) + E_{I_a, \hat{I}^t_s}\big(\text{log}(1 - D_s(\hat{h}_s^{t,i}) ) \big)},
  \end{aligned}
\end{equation}
where $L_g^t$ and $L_s^t$ provide effective constraints to guide the generator to achieve high-fidelity and semantic consistency during the process of cross-view translation. The total adversarial loss of PanoGAN is obtained by averaging the adversarial losses in all iterations, i.e.,
\begin{equation}
  L_{adv} = \frac{1}{T}\sum_{t=1}^{T} (L_g^t + L_s^t).
\end{equation}

\noindent \textbf{Pixel-Semantic Alignment Loss}. With the multi-scale feature alignment score map, a semantic alignment loss is computed to indicate how much content information of an aerial image can be translated to the ground-level domain. Specifically, the image semantic alignment loss $\tilde{L}_{g}^t$ and the segmentation semantic alignment loss $\tilde{L}_{s}^t$ at iteration loop $t$ can be expressed as
\begin{equation}
\begin{aligned}
  \tilde{L}_{g}^t & =\sum_{i=1}^{r}{E_{I_a, I_g}\big(\text{log}D_g(S_g^{t,i}) \big) + E_{I_a, \hat{I}^t_g}\big(\text{log}(1 - D_g(\hat{S}_g^{t,i}) ) \big)}, \\
  \tilde{L}_{s}^t & =\sum_{i=1}^{r}{E_{I_a, I_s}\big(\text{log}D_s(S_s^{t,i}) \big) + E_{I_a, \hat{I}^t_s}\big(\text{log}(1 - D_s(\hat{S}_s^{t,i}) ) \big)}.
  \end{aligned}
\end{equation}
By taking all the iterations into account, the total pixel-semantic alignment loss of PanoGAN is given by
\begin{equation}
  L_{fa} = \frac{1}{T}\sum_{t=1}^{T} ( \tilde{L}_{g}^t + \tilde{L}_{s}^t).
\end{equation}

\noindent \textbf{Reconstruction Loss}. We also use the $L1$ distance loss between the ground-truth training sample and the generated sample to compute the pixel-wise reconstruction loss to penalize the difference of the generated sample from the real sample.
The average reconstruction loss w.r.t panorama images and segmentation maps in the training stage can be formulated as
\begin{equation}
  L_{re} = \frac{1}{T}\sum_{t=1}^{T}\Big(E_{\hat{I}^t_g}\big( \|I_g-\hat{I}^t_g\|_1 \big) +
  E_{\hat{I}^t_s}\big( \|I_s-\hat{I}^t_s\|_1 \big) \Big).
\end{equation}

\noindent \textbf{Overall Loss.} The overall loss of PanoGAN is a weighted sum of the adversarial loss, pixel-semantic alignment loss, and reconstruction loss, so that
the generator $G$, discriminators $D_g$ and $D_s$ are trained alternatively in an end-to-end fashion to solve the following min-max optimizing problem,
\begin{equation}
	 G^{*}, D^{*}_g, D^{*}_s = \mathop{\arg} \mathop{\min}_{G} \mathop{\max}_{D_g, D_s}  (L_{adv} + L_{fa} + L_{re}).
\end{equation}

\subsection{Implementation Details}
To address the difference of image size between the aerial image and panorama image, we propose an image pre-processing strategy to make the input image have the same size as the output image. Consider the input aerial image $I_a$ is in size $m{\times} n$ and the panorama image $I_g$ is in size $p{\times} q$. The pre-processing procedure is comprised of three steps.
In the first step, the aerial image $I_a$ is re-scaled to $p {\times} \frac{q}{4}$. Then, we extend the aerial image to $p {\times} q$ by sequential rotation and multiplication. The first quarter patch of the new input image is the aerial image, the following quarter image patch is the 90-degree counter-clock rotation of the former quarter image patch. This process is repeated 3 times so that we can construct the final input image for cross-view panorama image generation.

There are the same number of encoding layers, decoding layers, and AFM layers in PanoGAN, which is set s $8$ in our experiments.
The image discriminator and segmentation discriminator adapt the multi-scale pyramid structure proposed in \cite{Liu_nips19}.
Instance normalization \cite{ulyanov2016instance} is adopted for the generator and the discriminators. 

PanoGAN is optimized with the Adam solver \cite{kingma2014adam} with the momentum values 0.9 and 0.999.
In the experiments, the feedback weighting parameters in Eq.~\eqref{eq:fbl} are set to  $\alpha_i=0.5 (i=1,2,3,4,5)$.
The number of feedback iterations is set to $2$.
The learning rate of the discriminator and generator is set to 0.00001 and 0.0001, respectively.

%------------------- Section 4: Implementation ----------------------------------------------------------------------------------
\section{Experiments}
\label{experiment}
%-------------------FIG: qualitative comparison on dataset CVUSA -----------------------------------------------------------
\begin{figure*}[!t] \small
    \centering
    \includegraphics[width=1\linewidth]{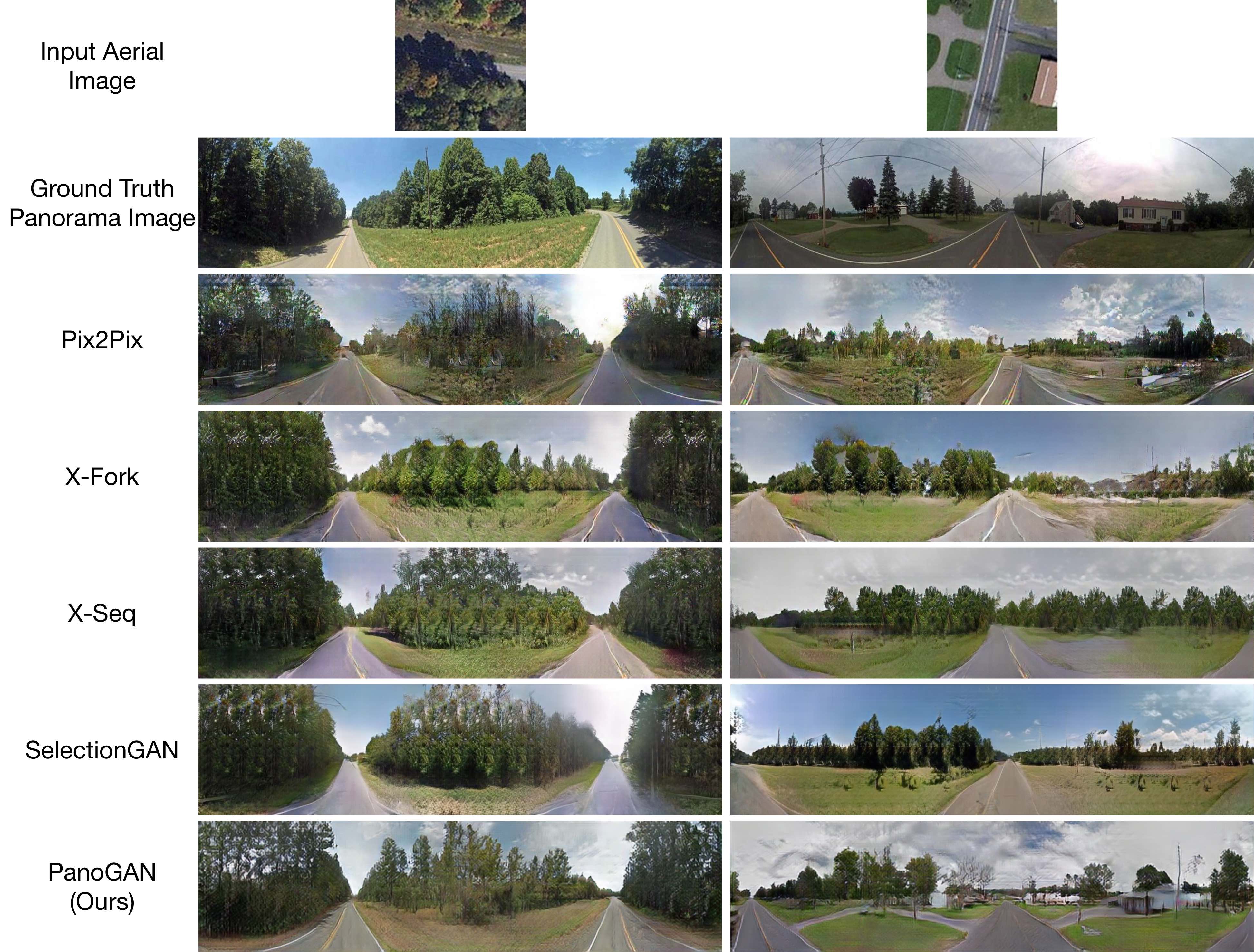}
    \caption{Qualitative comparison of different methods on the CVUSA dataset. From top to bottom: Input, Ground Truth, Pix2pix~\cite{isola2017image}, X-Fork~\cite{Krishna_cvpr18}, X-Seq~\cite{Krishna_cvpr18}, SelectionGAN~\cite{tang_cvpr19}, and PanoGAN (Ours).
    The proposed PanoGAN generates more realistic results with more convincing details than the leading methods on rural scenes.}
    \label{fig:quality_CVUSA}
\end{figure*}
%-------------------FIG End: qualitative comparison on dataset CVUSA -----------------------------------------------------------
%%-------------------TABLE: quantitative comparison on dataset CVUSA -----------------------------------------------------------=
\begin{table*}[!t] \small
\centering
\caption{Quantitative evaluation of the CVUSA dataset for cross-view panorama image generation. For all metrics except KL score, higher is better.
($\ast$) Inception Score for real (ground truth) data is 5.0508, 3.5041, and 5.1280 for all, top-1, and top-5 setups, respectively.
}
\resizebox{1\linewidth}{!}{%
  \begin{tabular}{lcccccccccccc} \toprule
   \multirow{2}{*}{Method} & \multicolumn{3}{c}{Inception Score$^\ast$ $\uparrow$} & \multicolumn{4}{c}{Accuracy (\%) $\uparrow$}  & \multirow{2}{*}{KL $\downarrow$} & \multirow{2}{*}{SSIM $\uparrow$} & \multirow{2}{*}{PSNR $\uparrow$} & \multirow{2}{*}{SD $\uparrow$}   \\
   \cmidrule(lr){2-4} \cmidrule(lr){5-8} & All & Top-1 & Top-5 & Top-1 (All) & Top-5 (All) & Top-1 (0.5) & Top-5 (0.5) \\ \hline
   Pix2pix \cite{isola2017image}              &    3.10        &    2.40         &  3.23
                        &    21.34     &     53.17    &    29.42    &  67.51
                        & 10.77$\pm$1.74
                       &     0.3658    &     18.4933    &     17.6953         \\

   X-Fork \cite{Krishna_cvpr18}               &    3.51        &    \textbf{2.79}         & 3.54
   &    29.01     &     49.32    &    44.04    &  76.13
                       & 6.60$\pm$1.55
                       &     0.4161    &     19.3859    &     18.3579         \\

   X-Seq \cite{Krishna_cvpr18}            &    2.86        &    2.58         &      2.95
                     &    26.94 &     58.88    &    39.29    &  71.89
                       & 7.61$\pm$1.71
                       &     0.4265    &     19.3464    &     18.4359        \\

   SelectionGAN \cite{tang_cvpr19}       &    3.29        &    2.65         &      3.32
   &    28.29     &     60.40    &    43.24    &  75.96
                        & 6.46$\pm$1.48
                       &     0.4315        &     20.2797        &     \textbf{19.1131}        \\

   PanoGAN (Ours)              &    \textbf{3.58}        &    2.75         &      \textbf{3.63}
                        &    \textbf{36.50}        &     \textbf{68.54}        &    \textbf{56.43}        &  \textbf{84.94}
                        & \textbf{4.20$\pm$1.19}
                       &     \textbf{0.4437}        &     \textbf{20.9467}    &     19.0913          \\
   \bottomrule
\end{tabular}}
\label{tab:cvusa}
\end{table*}
%%-------------------TABLE End: quantitative comparison on dataset CVUSA -----------------------------------------------------------=

\subsection{Experimental Setting}
\noindent \textbf{Datasets.}
We conduct experiments on two challenging datasets, i.e., the CVUSA dataset \cite{workman2015wide} and the Orlando-Pittsburgh (OP) dataset~\cite{regmi2019bridging}.
Both datasets are widely used for cross-view image translation and matching tasks.
Specifically, the CVUSA dataset contains 1.5 million geo-tagged pairs of ground and aerial images covering rural areas.
According to previous work \cite{Zhai_cvpr17,Krishna_cvpr18,tang_cvpr19}, we select a total 44,416 aerial-panorama image pairs in our experiments, which are further split into a training set of 35,532 image pairs and a test set of 8,884 image pairs.
The OP dataset covers urban area images of cities Orlando and Pittsburgh. It contains 1,910 training and 722 evaluation image pairs, each one consisting of a ground-level panorama and an aerial image.
For both datasets, aerial and panorama images are re-scaled to $256 {\times} 256$ and $256 {\times} 1024$, respectively.

%-------------------------------------FIGURE: qualitative comparison on dataset OP ------------------------------------------------------
\begin{figure*}[!t] \small
    \centering
    \includegraphics[width=1\linewidth]{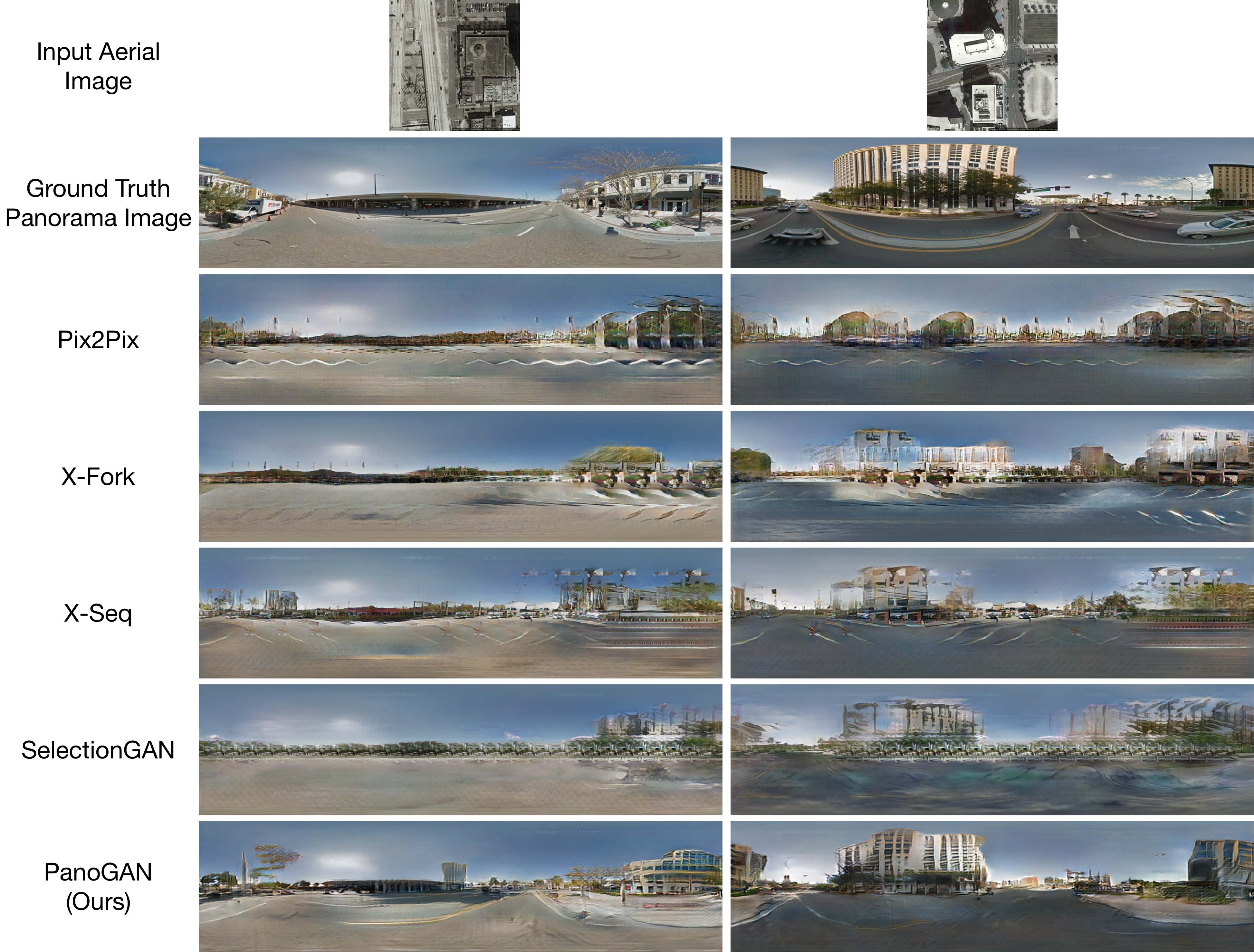}
    \caption{Qualitative comparison of different methods on the OP dataset. From top to bottom: Input, Ground Truth, Pix2pix~\cite{isola2017image}, X-Fork~\cite{Krishna_cvpr18}, X-Seq~\cite{Krishna_cvpr18}, SelectionGAN~\cite{tang_cvpr19}, and PanoGAN (Ours).
    The proposed PanoGAN generates more realistic results with more convincing details than the leading methods on urban scenes.}
    \label{fig:quality_op}
\end{figure*}
%-------------------------------------FIGURE End: qualitative comparison on dataset OP ------------------------------------------------------
%%-------------------TABLE: quantitative comparison on dataset OP -----------------------------------------------------------
\begin{table*}[!t] \small
\centering
\caption{Quantitative evaluation of the OP dataset for cross-view panorama image generation. For all metrics except KL score, higher is better.
($\ast$) Inception Score for real (ground truth) data is 3.4022, 2.5304, and 3.8007 for all, top-1, and top-5 setups, respectively.
}
\resizebox{1\linewidth}{!}{%
 \begin{tabular}{lcccccccccccc} \toprule
   \multirow{2}{*}{Method} & \multicolumn{3}{c}{Inception Score$^\ast$ $\uparrow$} & \multicolumn{4}{c}{Accuracy (\%) $\uparrow$}  & \multirow{2}{*}{KL $\downarrow$} & \multirow{2}{*}{SSIM $\uparrow$} & \multirow{2}{*}{PSNR $\uparrow$} & \multirow{2}{*}{SD $\uparrow$}   \\
   \cmidrule(lr){2-4} \cmidrule(lr){5-8} & All & Top-1 & Top-5 & Top-1 (All) & Top-5 (All) & Top-1 (0.5) & Top-5 (0.5) \\ \hline
   Pix2pix \cite{isola2017image}              &    2.1769        &    1.7148         &  2.3608
                        &    4.16     &     20.91    &    1.11     &  14.44
                      & 10.77$\pm$1.74
                       &     0.3658    &     18.4933    &     17.6953        \\

   X-Fork \cite{Krishna_cvpr18}             &    \textbf{2.3815}     &      \textbf{1.8835}    &     2.5536
                        &    0.55     &     2.49        &    2.22         & 22.22
                       & 14.83$\pm$1.74
                       &     0.4296    &     \textbf{19.2110}    & 17.4841     \\

   X-Seq \cite{Krishna_cvpr18}             &    2.1701        &    1.8032         &      2.3020
                        &    6.09 &     26.04    &    5.56    &  27.78
                       & 13.04$\pm$1.36
                       &     0.4311    &     19.1217    &     17.4841        \\

   SelectionGAN \cite{tang_cvpr19}      &    2.1119        &    1.8696         &      2.2050
                    &    6.09     &     31.16    &    3.33        &  31.11
                       & 12.75$\pm$1.32
                       &     0.4291    &     19.0980    &     17.9970         \\

   PanoGAN (Ours)         &    2.2331        &    1.8553         &      \textbf{2.5878}
                        &    \textbf{7.06}         &     \textbf{31.99}        &    \textbf{6.67}        &  \textbf{35.56}
                       & \textbf{7.89$\pm$1.12}
                       &     \textbf{0.4428}        &     19.0421        &     \textbf{18.0543}           \\
   \bottomrule
\end{tabular}}
\label{tab:op}
\end{table*}
%%-------------------TABLE End: quantitative comparison on dataset OP -----------------------------------------------------------

\noindent \textbf{Evaluation Metrics.}
We use Prediction Accuracy, Inception Score, KL score, Structural-Similarity (SSIM), Peak Signal-to-Noise Ratio (PSNR), and Sharpness Difference (SD) as the evaluation metrics, which are widely adopted by cross-view image synthesis methods \cite{Krishna_cvpr18,tang_cvpr19}. Specifically, the first three metrics measure the generated images in a high-level feature space, while the latter three metrics compare the fake and real images at a pixel similarity level.
\begin{itemize}[leftmargin=*]
\item \textbf{Prediction Accuracy \cite{Krishna_cvpr18}} measures the difference in class label prediction between real images and generated images with the same pre-trained AlexNet model as for computing Inception Score. Note that the prediction accuracy can be computed based on (i) all real samples or (ii) the real samples whose top-1 prediction exceeds~50\%.
\item \textbf{Inception Score \cite{salimans2016improved}} measures both the quality and diversity of synthesized images with the perspective of image classification. A higher Inception score indicates better image quality. In this paper, we follow \cite{Krishna_cvpr18} to compute the all, top-1, and top-5 inception scores.
\item \textbf{KL Score} \cite{nguyen2017dual} measures the distribution difference between the real and the synthesized image sets with KL divergence. The KL score is computed based on image features extracted by the pre-trained AlexNet model, which is expected to be a smaller value for better-synthesized images.
\item \textbf{Structural-Similarity (SSIM)} \cite{Zhou_Wang_tip04} measures the diversity of the synthesized samples. SSIM value ranges from 0.0 to 1.0 with higher SSIM values corresponding to more similar images in perception.
\item \textbf{Peak Signal-to-Noise Ratio (PSNR)} is computed as the ratio of the peak intensity value of a reference image to the root mean square reconstruction error (computed between image magnitudes) relative to the reference. A higher PSNR value indicates better image quality.
\item \textbf{Sharpness Difference (SD)} measures the similarity between real and generated images in the sharpness, which is computed based on the different gradients between real and generated images. A higher SD value indicates better image quality.
\end{itemize}

%-------------------Fig: ablation study -----------------------------------------------------------
\begin{figure*}[!t] \small
    \centering
    \includegraphics[width=1\linewidth]{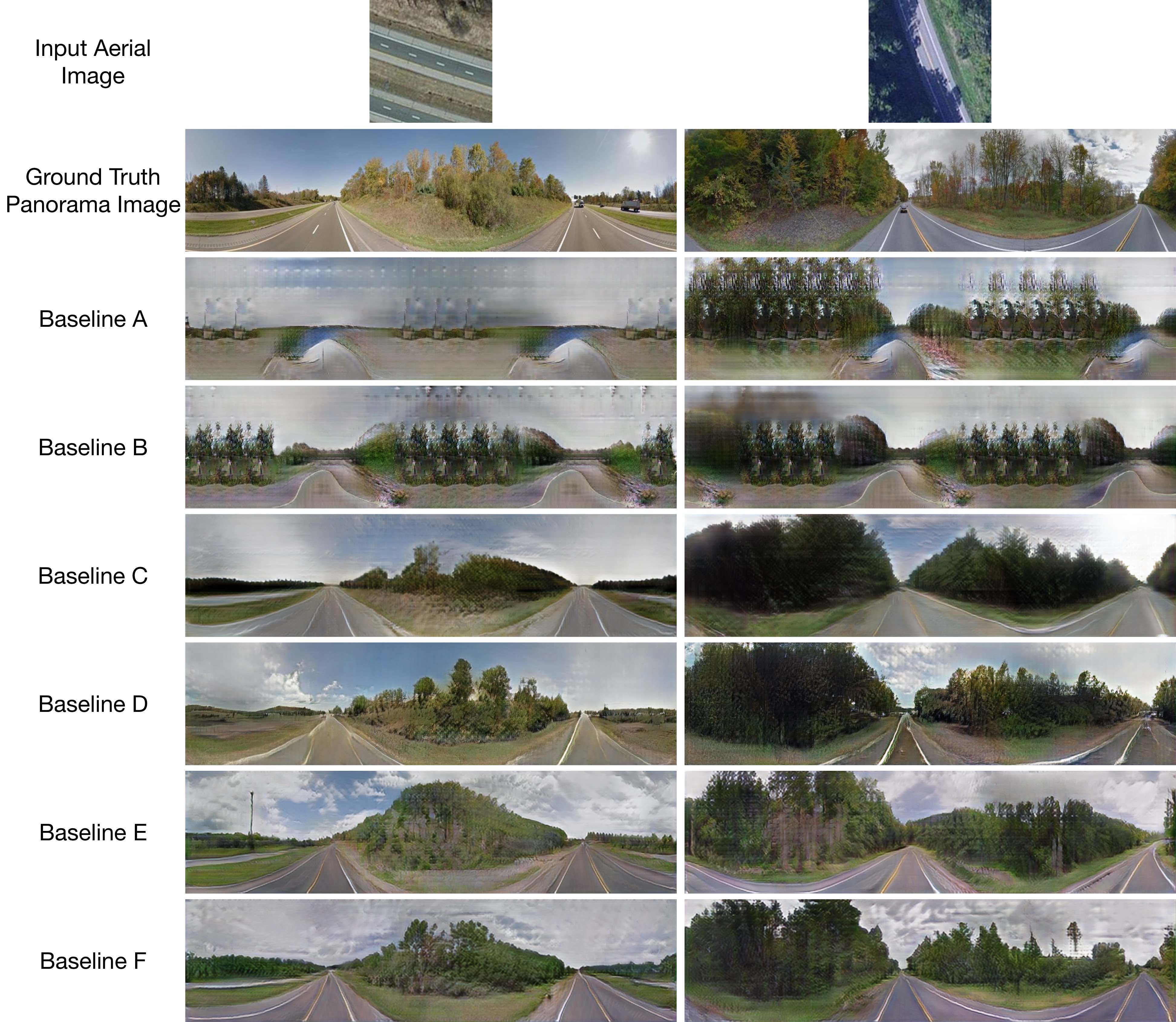}
    \caption{Qualitative comparison of different variants of the proposed PanoGAN on the CVUSA dataset. From top to bottom: Input, Ground Truth, Baseline A-F.}
    \label{fig:ablation}
    %\vspace{-0.4cm}
\end{figure*}
%-------------------Fig End: ablation study -----------------------------------------------------------
%-------------------Tab: ablation study -----------------------------------------------------------
\begin{table*}[!t] \small
\centering
\caption{Ablation study of the proposed PanoGAN on the CVUSA dataset.}
% \resizebox{1\linewidth}{!}{%
  \begin{tabular}{ccccccccccc} \toprule
   \multirow{2}{*}{Baseline} & \multirow{2}{*}{Setting} & \multicolumn{4}{c}{Accuracy (\%) $\uparrow$} & \multirow{2}{*}{KL$\downarrow$}  \\
   \cmidrule(lr){3-6}
   & & Top-1 (All) & Top-5 (All) & Top-1 (0.5) & Top-5 (0.5) \\ \hline
    A  &   U-Net + PatchGAN $\rightarrow$ Panorama Image    &    10.74     &     29.91    &    19.37    &  43.37
                       %&    2.8009        &    1.8249         &      2.7897
                       & 45.02$\pm$1.78 \\

    B   &  A + Dual-Branch Generation   &    24.01     &     53.55    &    36.09    &  68.18
                       %&    \textbf{3.4552}        &    \textbf{2.6597}         &      \textbf{3.6042}
                       & 8.99$\pm$1.62 \\

    C   &   B + Adversarial  Feedback  Module (AFM)   &    27.75     &     59.82    &    43.81    &  75.19
                       %&    2.8476        &    2.3020         &      2.8296
                       & 6.97$\pm$1.54 \\

    D   &  B + Our Discrimination Mechanism    &    28.15     &     60.43    &    42.90    &  74.39
                       %&    3.3470        &    2.5332         &      3.3953
                       &     6.87$\pm$1.55 \\

    E   &   C + Pyramid Discriminator \cite{Liu_nips19}    &    29.41         &     60.02        &    45.04        &  75.73
                       %&    2.9972        &    2.3603         &      3.0207
                       &  6.82$\pm$1.53     \\

    F   &   C + Our Discrimination Mechanism  &    \textbf{30.13}     &     \textbf{61.48}    &    \textbf{46.81}    &  \textbf{77.39}
                       %&    3.0072        &    2.3150        &          3.0798
                       & \textbf{5.89$\pm$1.35} \\
   \bottomrule
\end{tabular}
\label{tab:ablation}
\vspace{-0.4cm}
\end{table*}
%-------------------Tab End: ablation study -----------------------------------------------------------

\subsection{State-of-the-Art Comparison}
We compare the proposed PanoGAN with existent cross-view image generation methods, Pix2Pix~\cite{isola2017image}, X-Fork~\cite{Krishna_cvpr18}, X-Seq~\cite{Krishna_cvpr18}, and SelectionGAN~\cite{tang_cvpr19}.

As our task is to translate an aerial image to its corresponding panorama image, the aerial image is the only input for all involved methods in our experiments. Meanwhile, three types of ground-truth data, i.e., aerial image, panorama image, and panorama segmentation map, are provided for model training. To facilitate understanding, we summarize all the competitors' interface information in Table~\ref{tab:interface}. Specifically, Pix2pix translates aerial images to panorama images, thus it does not use ground-truth panorama images for model training. The other comparison methods translate aerial images into both panorama images and panorama segmentation maps, which means that their model can be trained with the semantic guidance of panorama segmentation maps.

We note that the original network architecture of SelectionGAN in \cite{tang_cvpr19} is designed to take input the concatenation of top-view image and ground-view segmentation map along channel dimension. Therefore, we adapt their models to our experimental setting by simply changing the input channel number from $6$ to $3$, i.e., the model input is replaced with $I_a$. We also note that \cite{Krishna_cvpr18} and \cite{tang_cvpr19} generate the first quarter of the ground level panorama image with size $256\times256$ in their experiments, which is different from generating 360-degree panorama images with size $256\times1024$ in our experiment. Thus, we need to retrain their models with panorama data. In the training phase, the hyper-parameters of the methods are set according to their papers. By following the experimental setting for model training in \cite{tang_cvpr19,regmi2019bridging}, the training epoch number is set as $30$ on the dataset CVUSA and $200$ on the dataset OP.

\noindent \textbf{Quantitative Evaluation.}
The quantitative results on CVUSA are presented in Table~\ref{tab:cvusa}. Note that PanoGAN performs the best on most metrics. In terms of high-level image evaluation metrics, PanoGAN achieves the best results on all metrics except for the Top-1 Inception score. In particular, PanoGAN yields notable improvements in prediction accuracy and KL score. This demonstrates that PanoGAN can extract image semantic information from aerial images and successfully transfer it to the corresponding panorama images. In terms of low-level image evaluation metrics, PanoGAN outperforms other methods on SSIM and PSNR, and achieves the second-best result on SD. These results indicate that PanoGAN can produce panorama images with higher image quality than existing leading models.

Table~\ref{tab:op} shows the quantitative comparison on the OP dataset.
We observe that PanoGAN achieves better results than the leading methods on most high-level and low-level evaluation metrics.
Note that the training samples are limited in the OP dataset, which brings considerable challenges to cross-view panorama image generation.

\noindent \textbf{Qualitative Evaluation.}
Fig.~\ref{fig:quality_CVUSA} shows the qualitative results compared with existing methods on CVUSA.
We can see that the proposed PanoGAN generates more photo-realistic panorama images with more similarity in semantics to the ground truth panorama images than other leading methods. Concretely, in the relatively simple scenario shown in the first column, all the methods capture the layout characteristics of the location and transfer the information into panorama images.
We also note that PanoGAN produces a better panorama image with fewer visual artifacts (e.g., plants and vegetation) and more object details (e.g., road and street lines).
In the relatively complex scenario with more objects in the location, PanoGAN significantly outperforms other methods by accurately preserving the image layout.
For example, in the second column of Fig.~\ref{fig:quality_CVUSA}, PanoGAN successfully generates the path on one side of the road and the house on the other size, while other methods fail to generate the objects and details.

Fig.~\ref{fig:quality_op} presents the urban panorama image generation on OP.
We see that our PanoGAN generates plausible objects in the urban panorama images such as overpasses, buildings, and roads, while the comparing methods including the state-of-the-art SelectionGAN nearly fail in this task.
Moreover, we observe that PanoGAN performs worse on the OP dataset than on the CVUSA dataset by introducing more artifacts.
There are two reasons: (i) the facades of buildings in urban panorama contain much more complex textures than the suburb panorama images increasing the difficulty of cross-view panorama image generation; (ii) the available 1,910 training samples in OP cannot represent the image distribution. Thus, it is extremely challenging to learn the mapping from top-view to ground-view.

%-------------------Fig: Input format -----------------------------------------------------------
\begin{figure*}[!t] \small
    \centering
    \includegraphics[width=1\linewidth]{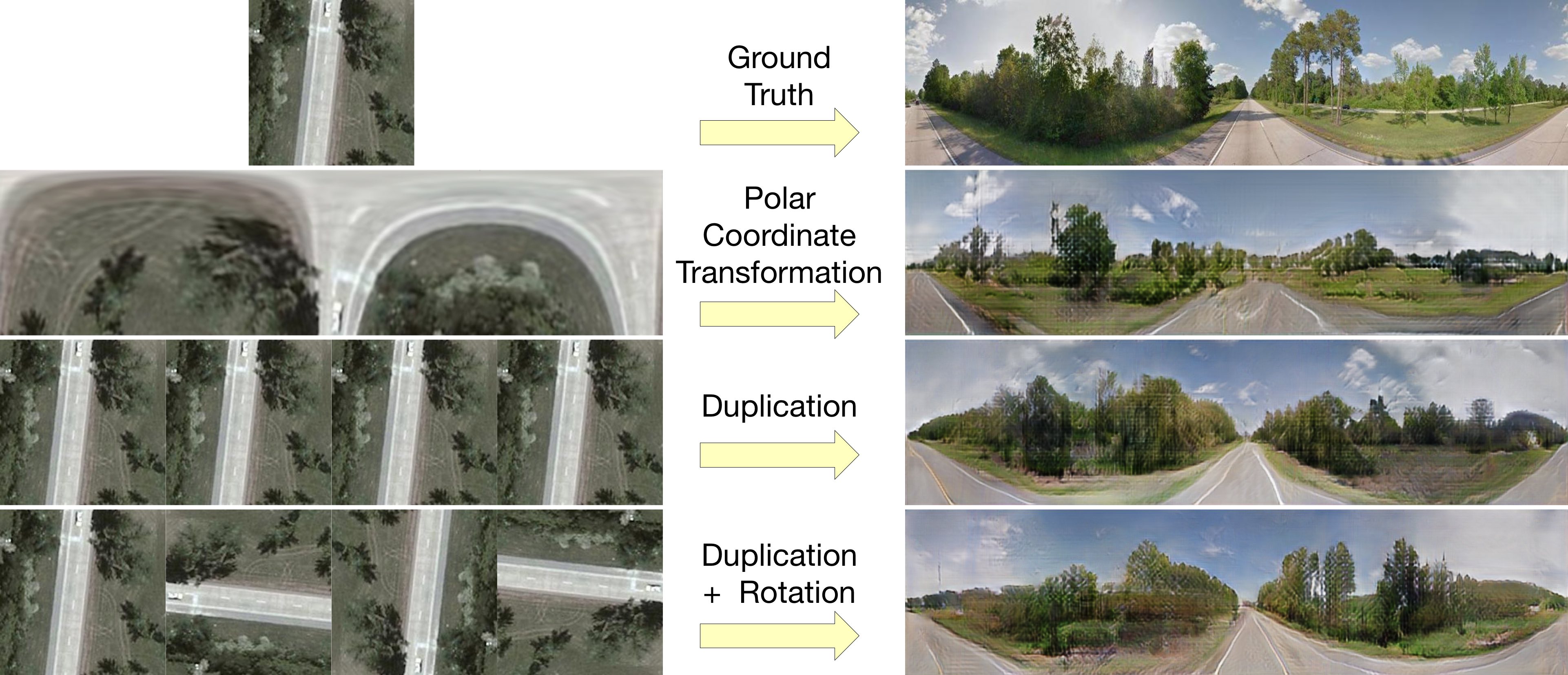}
    \caption{Qualitative comparison of the proposed PanoGAN with different input image format on the CVUSA dataset. From top to bottom: Input, Ground Truth, Polar Coordinate  transformation, Duplication without rotation, and Duplication with rotation.}
    \label{fig:input_preprocess}
    %\vspace{-0.4cm}
\end{figure*}
%-------------------Fig End: Input format -----------------------------------------------------------
%-------------------TABLE: Input format -----------------------------------------------------------
\begin{table*}[htp] \small
\centering
\caption{Quantitative evaluation of different input formats for cross-view panorama image generation. For all metrics except KL score, higher is better.}
\resizebox{1\linewidth}{!}{
  \begin{tabular}{lccccccccc}
  \toprule
  \multirow{2}{*}{Input Format} & \multicolumn{4}{c}{Accuracy (\%) $\uparrow$} & \multirow{2}{*}{KL $\downarrow$}  & \multirow{2}{*}{SSIM $\uparrow$} & \multirow{2}{*}{PSNR $\uparrow$} & \multirow{2}{*}{SD $\uparrow$} \\
\cmidrule(lr){2-5}
                          & Top-1 (All)  & Top-5 (All) & Top-1 (0.5)       & Top-5 (0.5) \\
\hline
   Polar Coordinate Transformation              &  18.75  &  48.36  &  27.48  &  60.60  &  15.19$\pm$1.62  &  0.4128  &  19.1060  &  18.7183    \\
   Duplication                &  30.05  &  61.22  &  46.64  &  76.86  &  5.94$\pm$1.35  &  0.4098  &  19.1291  &  18.6309    \\
   Duplication$+$Rotation          &  30.13  &  61.48  &  46.81  &  77.40  &  5.85$\pm$1.30  &  0.4198  &  19.2712  &  18.7535    \\
\bottomrule
\end{tabular}}
\label{tab:input_preprocess}
%\vspace{-0.4cm}
\end{table*}
%-------------------TABLE End: Input format -----------------------------------------------------------\

\subsection{Ablation Study}
We also conduct extensive ablation studies on CVUSA to evaluate the effectiveness of each component of the proposed PanoGAN.
To save training time, PanoGAN is trained with randomly selected $8,883$ training samples which are nearly $1/4$ of all the training samples and evaluated on the test samples.
We consider six baseline models for PanoGAN:
\begin{itemize}[leftmargin=*]
  \item \textbf{Baseline A} uses an encoder-decoder architecture to generate panorama images from aerial images, in which there is a U-Net based generator and a PatchGAN based discriminator.
  \item \textbf{Baseline B (A + Dual-Branch Generation)} generates a panorama image and segmentation map by forcing the output of the baseline A to be 6-channel. In addition, there are two discriminators in baseline~B correspond to the panorama image and segmentation map, respectively.
  \item \textbf{Baseline C (B + AFM)} adds the proposed Adversarial Feedback Module (AFM) to baseline B.
  \item \textbf{Baseline D (B + Our Discrimination Mechanism)} replaces the PatchGAN based discriminators in baseline B with our proposed discrimination mechanism.
  \item \textbf{Baseline E (C + Pyramid Discriminator~\cite{Liu_nips19})} employs the pyramid discriminator proposed in \cite{Liu_nips19} for discrimination on the basis of baseline C.
  \item \textbf{Baseline F (C + Our Discrimination Mechanism)} is the full PanoGAN model that uses the proposed discrimination mechanism on the basis of baseline C.
\end{itemize}

\noindent \textbf{Effectiveness of Dual-Branch Generation.}
The results of the ablation study are shown in Fig.~\ref{fig:ablation} and Table \ref{tab:ablation}.
We observe that baseline B remarkably outperforms baseline A on all evaluation metrics,
which means that it is effective to generate semantically consistent images with the aid of segmentation maps in the model training process. The results also confirm our motivation for the dual-branch generation strategy.

\noindent \textbf{Effectiveness of Adversarial Feedback Learning.}
We note that Baseline C significantly outperforms Baseline B on all evaluation metrics.
Specifically, the proposed adversarial feedback module achieves around 3.7 and 7.7 point gains on top-1 accuracy, 6.4 and 7.0 point gains on top-5 accuracy and 2.0 point gains on KL score.
These results confirm the effectiveness of the proposed adversarial feedback module and the proposed feedback layer, both playing an essential role in generating high-quality panorama images.

The effectiveness of the proposed AFM can be further validated by comparing baseline F with baseline D.
Baseline F further boosts the performance on most metrics by adding AFM to baseline D.

\noindent \textbf{Effectiveness of Pixel-Semantic Alignment Discriminator.}
Compared with baseline B, baseline D achieves significantly better performance, which verifies the effectiveness of the proposed joint semantic-image crossing embedding discriminators.

Besides, by comparing the performance of baseline F with that of baseline E, we observe that around 0.7 and 1.7 point gains on Top-1 accuracy, 1.4 and 1.6 gain on Top-5 accuracy, and 1.0 point gains on KL score are obtained.
These improvements demonstrate the advantage of our proposed discrimination compared with the recent advanced pyramid multi-scale feature discriminator.

\subsection{Model Analysis of PanoGAN}
\noindent \textbf{Influence of Different Input Formats.}
We investigate the influence of three different input formats to the performance of our PanoGAN, as shown in Table~\ref{tab:input_preprocess}.
The first method is `Polar Coordinate Transformation', which transforms the $256 {\times} 256$ aerial image from the Cartesian coordinate system to the Polar coordinate system for producing $256 {\times} 1024$ unfolded image as input. The second method is to concatenate duplicated aerial images along with the width axis without image rotation, which is denoted by `Duplication'.
The last pre-processing method is denoted by `Duplication+Rotation', which concatenates duplicated aerial images along with the width axis with image rotation.
All three pre-processing strategies are evaluated on CVUSA.

The comparison results are shown in Table~\ref{tab:input_preprocess} and Fig.~\ref{fig:input_preprocess}.
We observe that the proposed `Duplication+Rotation' strategy achieves the best results on all metrics.
We can obtain the similar conclusion from Fig.~\ref{fig:input_preprocess}.
Thus, we adopt a 'rotation-duplication' strategy in our experiments, which alleviates the different size problems in the challenging cross-view panorama generation task.
%-------------------Fig: Feedback Loop in the test phase -----------------------------------------------------------
\begin{figure}[!t] \small
    \centering
    \includegraphics[width=1\linewidth]{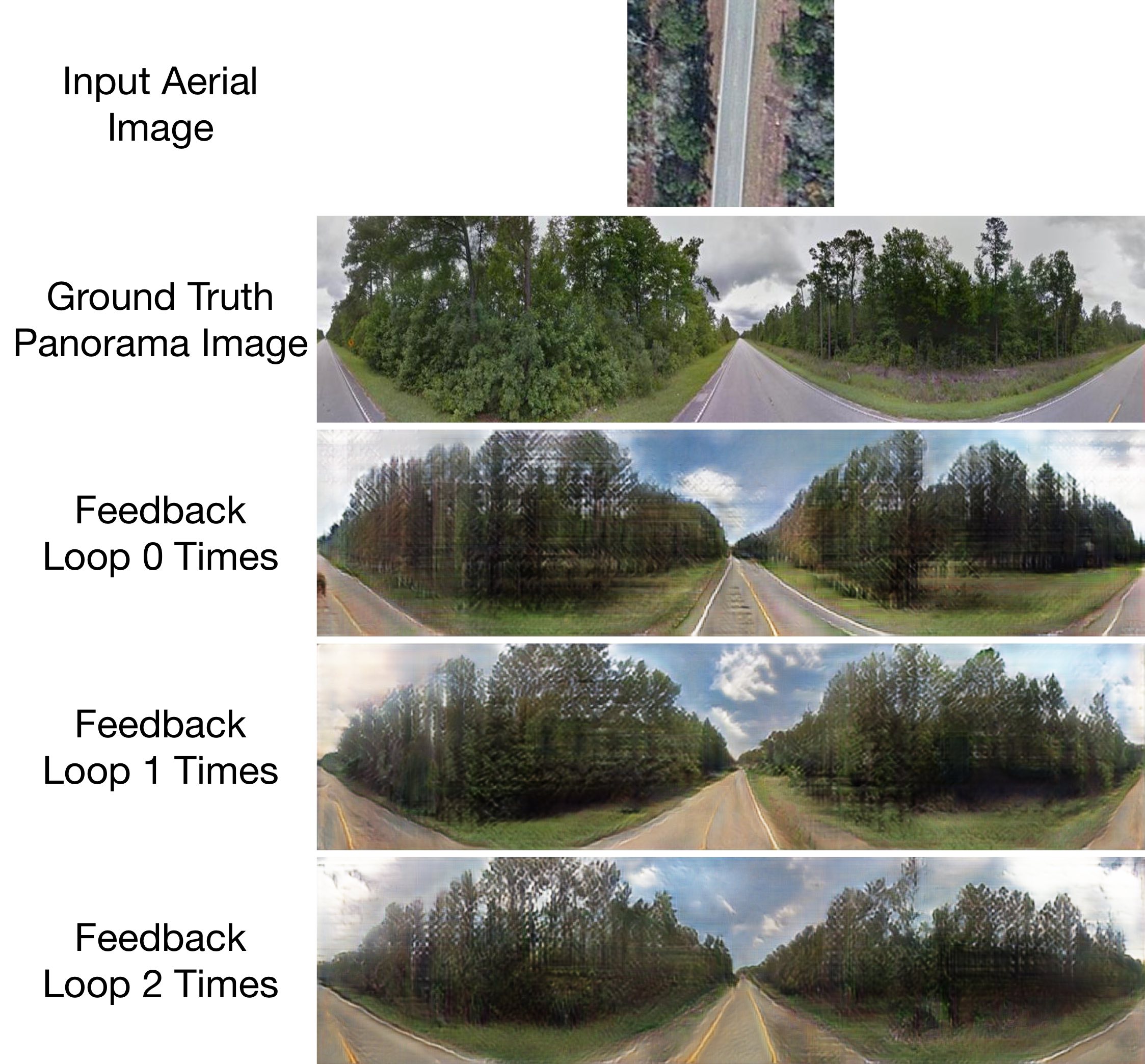}
    \caption{Qualitative comparison of different numbers of feedback loops of the proposed PanoGAN on the CVUSA dataset. From top to bottom: Input, Ground Truth, Loop 0, 1, and 2 times.}
    \label{fig:feedback_loop_train}
    %\vspace{-0.4cm}
\end{figure}
%-------------------Fig: Feedback Loop in the test phase -----------------------------------------------------------
%-------------------TAB: Feedback Loop in training phase -----------------------------------------------------------
\begin{table*}[htp] \small
\centering
\caption{The number of feedback loops v.s. quantitative evaluation on the CVUSA dataset for cross-view panorama image generation. For all metrics except KL score, higher is better.}
% \resizebox{1\linewidth}{!}{
  \begin{tabular}{cccccccccc}
  \toprule
  \multirow{2}{*}{\# Feedback Loop} & \multicolumn{4}{c}{Accuracy (\%) $\uparrow$} & \multirow{2}{*}{KL $\downarrow$} & \multirow{2}{*}{SSIM $\uparrow$} & \multirow{2}{*}{PSNR $\uparrow$} & \multirow{2}{*}{SD $\uparrow$}  \\
\cmidrule(lr){2-5}
                          & Top-1 (All)  & Top-5 (All) & Top-1 (0.5)       & Top-5 (0.5) \\
\hline
   0   &  22.67  &  55.63  &  34.49  &  69.68  &  8.68$\pm$1.44 &  0.4017  &  19.2699  &  18.5640     \\
   1                 &  29.41  &  \textbf{62.03}  &  45.18  &  \textbf{77.90}  &  6.05$\pm$1.43   &  0.4165  &  19.2702  &  18.7199  \\
   2                 &  \textbf{30.13}  &  61.48  &  \textbf{46.81}  &  77.40  &  \textbf{5.85$\pm$1.31}  &  \textbf{0.4198}  &  \textbf{19.2712}  &  \textbf{18.7535}   \\
\bottomrule
\end{tabular}
\label{tab:feedbak_loop_train}
%\vspace{-0.4cm}
\end{table*}
%-------------------TAB End: Feedback Loop in training phase -----------------------------------------------------------
%-------------------TAB: Feedback Loop in the test phase -----------------------------------------------------------
\begin{table*}[htp] \small
\centering
\caption{The number of feedback loops in the inference stage v.s. quantitative evaluation on the CVUSA dataset for cross-view panorama image generation. For all metrics except KL score, the higher is better.}
 \begin{tabular}{cccccccccc}
  \toprule
  \multirow{2}{*}{\# Feedback Loop} & \multicolumn{4}{c}{Accuracy (\%) $\uparrow$} & \multirow{2}{*}{KL $\downarrow$}
  & \multirow{2}{*}{SSIM $\uparrow$} & \multirow{2}{*}{PSNR $\uparrow$} & \multirow{2}{*}{SD $\uparrow$}  \\
\cmidrule(lr){2-5} & Top-1 (All)  & Top-5 (All) & Top-1 (0.5)       & Top-5 (0.5) \\
\hline
   0                 &  11.93  &  37.30  &  18.90  &  48.38  &   42.51$\pm$2.11    &  0.4144   &  19.5731  &  18.6924   \\
   1                 &  35.55  &  68.30  &  55.26  &  84.91  &    4.47$\pm$1.20    &  0.4349   &  20.5131  &  18.8272    \\
   2                 &  \textbf{36.50}  &  \textbf{68.54}  &  \textbf{56.43}  &  \textbf{84.94}  &    \textbf{4.20$\pm$1.19}    &  \textbf{0.4437}   &  \textbf{20.9467}  &  \textbf{19.0913}     \\
   3                 &  34.83  &  67.24  &  54.32  &  84.31  &    4.45$\pm1.20$    &  0.4261   &  20.3173  &  18.5862    \\
   4                 &  33.19  &  65.13  &  50.95  &  82.24  &    4.84$\pm1.25$    &  0.4226   &  20.1650  &  18.5804     \\
\bottomrule
\end{tabular}
\label{tab:feedbak_loop_test}
%\vspace{-0.4cm}
\end{table*}
%-------------------TAB End: Feedback Loop in the test phase -----------------------------------------------------------

\noindent \textbf{Influence of the Number of Feedback Loops.}
The number of feedback loops influences the performance of PanoGAN both in model training and in inference stages.
In each mini-batch of the model training stage, the generation and discrimination losses computed at the end of each feedback loop are summed up for the network parameter update. The number of feedback loops determines the objective function optimization of PanoGAN.
In the model inference stage, PanoGAN generates panorama images with the fixed network parameter, but the quality of panorama images is affected by the number of feedback loops adopted during image generation.

We firstly explore the number of feedback loops in the training stage.
We set the number of the feedback loops as $k$ and set $k$ to 0, 1, 2, respectively.
Moreover, we set the number of feedback loops $j$ in the inference stage similar to $k$, i.e., $j{=}k$.
Note that $k{=}0$ indicates PanoGAN is just a forward network without any feedback loop.

We show the quantitative results in Table \ref{tab:feedbak_loop_train} and the qualitative results in Fig. \ref{fig:feedback_loop_train}.
We observe from Table~\ref{tab:feedbak_loop_train} that PanoGAN performs better on $k{>}0$ on all evaluation metrics, especially on high-level feature-based metrics, i.e., prediction accuracy and KL score. These results demonstrate that the proposed feedback learning module is beneficial to transfer high-level semantic information across different views.
We also observe that increasing the number of feedback loops promotes the performance on most metrics, but the improvement is limited.
Fig.~\ref{fig:feedback_loop_train} presents the consistent results that the panorama image generated with no feedback loop introduces obvious visual artifacts, such as trees and roads.
Thus, we set $k{=}2$ in our experiments.

Secondly, PanoGAN model is trained based on 35,532 samples of CVUSA with the feedback loop number being set as $k{=}2$, then is tested on the test set of CVUSA with various feedback loop numbers $j {\in} \{0,1,2,3,4\}$.
The goal is to investigate the effect of the feedback loop procedure in PanoGAN to the quality of the generated panorama image. From the quantitative comparison results presented in Table~\ref{tab:feedbak_loop_test}, we observe that PanoGAN performs poorly when $j{=}0$, indicating that the Adversarial feedback mechanism (AFM) is an indispensable component of the iterative generation framework of PanoGAN for high-quality panorama image synthesis. Besides, we observe that PanoGAN achieves the best performance when $j{=}k$, suggesting that the same feedback loop number should be adopted in both the training and test phase for PanoGAN generating panorama images in high quality.

%------------------- Section 5: Conclusion ----------------------------------------------------------------------------------
\section{Conclusion}
\label{conclusions}
We propose a novel Panorama GAN (PanoGAN) for generating panorama ground-view images from top-view images.
PanoGAN introduces three key components: a dual-branch generation strategy, a kind of pixel-semantic alignment discriminator, and an adversarial feedback module.
The first component is employed to simultaneously generate both the targeted panorama image and its segmentation map for generating semantically consistent results.
The second one is used to enhance the fine details and semantic alignment between the generated results and the real ones.
The third one is adopted to progressively improve the generation performance under the feedback of the discriminators.
Extensive experiments on two challenging datasets show that the proposed PanoGAN achieves significantly better results than state-of-the-art approaches.

%------------------- Section *: Conclusion ----------------------------------------------------------------------------------
\section*{Acknowledgments}
This work was supported in part by the National Natural Science Foundation of China under Grant 61933013 and 62176069, in part by the Natural Science Foundation of Guangdong Province under Grant No.2019A1515011076, in part by the Innovation Group of Guangdong Education Department under Grant No.2020KCXTD014, in part by the International Science and Technology Cooperation Project of Jiangsu Province under Grant No.BZ2020069, and in part by Major Program of University Natural Science Research of Jiangsu Province under Grant No.21KJA520001.

%\input{appendix}

%\clearpage

\bibliographystyle{IEEEtran}
\bibliography{IEEEabrv,panogan}

\begin{thebibliography}{10}
\providecommand{\url}[1]{#1}
\csname url@samestyle\endcsname
\providecommand{\newblock}{\relax}
\providecommand{\bibinfo}[2]{#2}
\providecommand{\BIBentrySTDinterwordspacing}{\spaceskip=0pt\relax}
\providecommand{\BIBentryALTinterwordstretchfactor}{4}
\providecommand{\BIBentryALTinterwordspacing}{\spaceskip=\fontdimen2\font plus
\BIBentryALTinterwordstretchfactor\fontdimen3\font minus
  \fontdimen4\font\relax}
\providecommand{\BIBforeignlanguage}[2]{{%
\expandafter\ifx\csname l@#1\endcsname\relax
\typeout{** WARNING: IEEEtran.bst: No hyphenation pattern has been}%
\typeout{** loaded for the language `#1'. Using the pattern for}%
\typeout{** the default language instead.}%
\else
\language=\csname l@#1\endcsname
\fi
#2}}
\providecommand{\BIBdecl}{\relax}
\BIBdecl

\bibitem{Zhai_cvpr17}
M.~Zhai, Z.~Bessinger, S.~Workman, and N.~Jacobs, ``Predicting ground-level
  scene layout from aerial imagery,'' in \emph{Computer Vision and Pattern
  Recognition (CVPR)}, 2017.

\bibitem{Yi_Zhou_bmvc17}
Y.~Zhou and L.~Shao, ``Cross-view gan based vehicle generation for
  re-identification,'' in \emph{British Machine Vision Conference (BMVC)},
  2017.

\bibitem{yang2015weakly}
J.~Yang, S.~E. Reed, M.-H. Yang, and H.~Lee, ``Weakly-supervised disentangling
  with recurrent transformations for 3d view synthesis,'' in \emph{Advances in
  Neural Information Processing Systems (NeurIPS)}, 2015.

\bibitem{Sixing_Hu_cvpr18}
S.~Hu, M.~Feng, R.~M.~H. Nguyen, and G.~H. Lee, ``Cvm-net: Cross-view matching
  network for image-based ground-to-aerial geo-localization,'' in
  \emph{Computer Vision and Pattern Recognition (CVPR)}, 2018.

\bibitem{regmi2019bridging}
K.~Regmi and M.~Shah, ``Bridging the domain gap for ground-to-aerial image
  matching,'' in \emph{International Conference on Computer Vision (ICCV)},
  2019.

\bibitem{Scott_Workman_iccv15}
S.~Workman, R.~Souvenir, and N.~Jacobs, ``Wide-area image geolocalization with
  aerial reference imagery,'' in \emph{International Conference on Computer
  Vision (ICCV)}, 2015.

\bibitem{tian2017cross}
Y.~Tian, C.~Chen, and M.~Shah, ``Cross-view image matching for geo-localization
  in urban environments,'' in \emph{Computer Vision and Pattern Recognition
  (CVPR)}, 2017.

\bibitem{Krishna_cvpr18}
K.~Regmi and A.~Borji, ``Cross-view image synthesis using conditional gans,''
  in \emph{Computer Vision and Pattern Recognition (CVPR)}, 2018.

\bibitem{tang_cvpr19}
H.~Tang, D.~Xu, N.~Sebe, Y.~Wang, J.~J. Corso, and Y.~Yan, ``Multi-channel
  attention selection gan with cascaded semantic guidance for cross-view image
  translation,'' in \emph{Computer Vision and Pattern Recognition (CVPR)},
  2019.

\bibitem{tang2019local}
H.~Tang, D.~Xu, Y.~Yan, P.~H. Torr, and N.~Sebe, ``Local class-specific and
  global image-level generative adversarial networks for semantic-guided scene
  generation,'' in \emph{Computer Vision and Pattern Recognition (CVPR)}, 2020.

\bibitem{workman2015wide}
S.~Workman, R.~Souvenir, and N.~Jacobs, ``Wide-area image geolocalization with
  aerial reference imagery,'' in \emph{International Conference on Computer
  Vision (ICCV)}, 2015.

\bibitem{WWang_tmm2020}
W.~Wang, X.~Alameda-Pineda, D.~Xu, E.~Ricci, and N.~Sebe, ``Learning how to
  smile: Expression video generation with conditional adversarial recurrent
  nets,'' \emph{IEEE Transactions on Multimedia}, vol.~22, no.~11, pp.
  2808--2819, 2020.

\bibitem{guo2019fusegan}
X.~Guo, R.~Nie, J.~Cao, D.~Zhou, L.~Mei, and K.~He, ``Fusegan: Learning to fuse
  multi-focus image via conditional generative adversarial network,''
  \emph{IEEE Transactions on Multimedia (TMM)}, vol.~21, no.~8, pp. 1982--1996,
  2019.

\bibitem{zhang_tpami2019_stackgan++}
H.~Zhang, T.~Xu, H.~Li, S.~Zhang, X.~Wang, X.~Huang, and D.~N. Metaxas,
  ``Stackgan++: Realistic image synthesis with stacked generative adversarial
  networks,'' \emph{IEEE transactions on pattern analysis and machine
  intelligence}, vol.~41, no.~8, pp. 1947--1962, 2019.

\bibitem{zhou2019branchgan}
Y.-F. Zhou, R.-H. Jiang, X.~Wu, J.-Y. He, S.~Weng, and Q.~Peng, ``Branchgan:
  Unsupervised mutual image-to-image transfer with a single encoder and dual
  decoders,'' \emph{IEEE Transactions on Multimedia (TMM)}, vol.~21, no.~12,
  pp. 3136--3149, 2019.

\bibitem{mirza2014conditional}
M.~Mirza and S.~Osindero, ``Conditional generative adversarial nets,''
  \emph{arXiv preprint arXiv:1411.1784}, 2014.

\bibitem{tang2020xinggan}
H.~Tang, S.~Bai, L.~Zhang, P.~H. Torr, and N.~Sebe, ``Xinggan for person image
  generation,'' in \emph{European Conference on Computer Vision (ECCV)}, 2020.

\bibitem{gu2020toward}
X.~Gu, J.~Yu, Y.~Wong, and M.~S. Kankanhalli, ``Toward multi-modal conditioned
  fashion image translation,'' \emph{IEEE Transactions on Multimedia (TMM)},
  2020.

\bibitem{wang2020learning}
W.~Wang, X.~Alameda-Pineda, D.~Xu, E.~Ricci, and N.~Sebe, ``Learning how to
  smile: Expression video generation with conditional adversarial recurrent
  nets,'' \emph{IEEE Transactions on Multimedia (TMM)}, 2020.

\bibitem{wu2019relgan}
P.-W. Wu, Y.-J. Lin, C.-H. Chang, E.~Y. Chang, and S.-W. Liao, ``Relgan:
  Multi-domain image-to-image translation via relative attributes,'' in
  \emph{International Conference on Computer Vision (ICCV)}, 2019.

\bibitem{zhang2017stackgan}
H.~Zhang, T.~Xu, H.~Li, S.~Zhang, X.~Wang, X.~Huang, and D.~N. Metaxas,
  ``Stackgan: Text to photo-realistic image synthesis with stacked generative
  adversarial networks,'' in \emph{International Conference on Computer Vision
  (ICCV)}, 2017.

\bibitem{isola2017image}
P.~Isola, J.-Y. Zhu, T.~Zhou, and A.~A. Efros, ``Image-to-image translation
  with conditional adversarial networks,'' in \emph{Computer Vision and Pattern
  Recognition (CVPR)}, 2017.

\bibitem{Dosovitskiy_tpami17}
A.~Dosovitskiy, J.~T. Springenberg, M.~Tatarchenko, and T.~Brox, ``Learning to
  generate chairs, tables and cars with convolutional networks,'' \emph{IEEE
  Transactions on Pattern Analysis and Machine Intelligence (TPAMI)}, vol.~39,
  no.~4, pp. 692--705, 2017.

\bibitem{li2018hole}
S.~Li, C.~Zhu, and M.-T. Sun, ``Hole filling with multiple reference views in
  dibr view synthesis,'' \emph{IEEE Transactions on Multimedia (TMM)}, vol.~20,
  no.~8, pp. 1948--1959, 2018.

\bibitem{Zhou_eccv16}
T.~Zhou, S.~Tulsiani, W.~Sun, J.~Malik, and A.~A. Efros, ``View synthesis by
  appearance flow,'' in \emph{European Conference on Computer Vision (ECCV)},
  2016.

\bibitem{liu2020exocentric}
G.~Liu, H.~Tang, H.~Latapie, and Y.~Yan, ``Exocentric to egocentric image
  generation via parallel generative adversarial network,'' in
  \emph{International Conference on Acoustics, Speech and Signal Processing
  (ICASSP)}, 2020.

\bibitem{Regmi_cvpr18}
K.~Regmi and A.~Borji, ``Cross-view image synthesis using conditional gans,''
  in \emph{Proceedings of the IEEE Conference on Computer Vision and Pattern
  Recognition}, 2018, pp. 3501--3510.

\bibitem{Sanath_Narayan_eccv20}
S.~Narayan, A.~Gupta, F.~S. Khan, C.~G.~M. Snoek, and L.~Shao, ``Latent
  embedding feedback and discriminative features for zero-shot
  classification,'' in \emph{European Conference on Computer Vision (ECCV)},
  2020.

\bibitem{Joao_Carreira_cvpr16}
J.~Carreira, P.~Agrawal, K.~Fragkiadaki, and J.~Malik, ``Human pose estimation
  with iterative error feedback,'' in \emph{Computer Vision and Pattern
  Recognition (CVPR)}, 2016.

\bibitem{Shama_iccv19}
F.~Shama, R.~Mechrez, A.~Shoshan, and L.~Zelnik-Manor, ``Adversarial feedback
  loop,'' in \emph{International Conference on Computer Vision (ICCV)}, 2019.

\bibitem{Huh_cvpr19}
M.~Huh, S.-H. Sun, and N.~Zhang, ``Feedback adversarial learning: Spatial
  feedback for improving generative adversarial networks,'' in \emph{Computer
  Vision and Pattern Recognition (CVPR)}, 2019.

\bibitem{Wang_cvpr18}
T.-C. Wang, M.-Y. Liu, J.-Y. Zhu, A.~Tao, J.~Kautz, and B.~Catanzaro,
  ``High-resolution image synthesis and semantic manipulation with conditional
  gans,'' in \emph{Computer Vision and Pattern Recognition (CVPR)}, 2018.

\bibitem{Lin_cvpr17}
G.~Lin, A.~Milan, C.~Shen, and I.~Reid, ``Refine{N}et: {M}ulti-path refinement
  networks for high-resolution semantic segmentation,'' in \emph{Computer
  Vision and Pattern Recognition (CVPR)}, 2017.

\bibitem{Long_cvpr15}
J.~Long, E.~Shelhamer, and T.~Darrell, ``Fully convolutional networks for
  semantic segmentation,'' in \emph{Computer Vision and Pattern Recognition
  (CVPR)}, 2015.

\bibitem{Kim_icml17}
T.~Kim, M.~Cha, H.~Kim, J.~K. Lee, and J.~Kim, ``Learning to discover
  cross-domain relations with generative adversarial networks,'' in
  \emph{International Conference on Machine Learning (ICML)}, 2017.

\bibitem{TLin_cvpr17}
T.-Y. Lin, P.~Doll{\'a}r, R.~Girshick, K.~He, B.~Hariharan, and S.~Belongie,
  ``Feature pyramid networks for object detection,'' in \emph{Computer Vision
  and Pattern Recognition (CVPR)}, 2017.

\bibitem{Liu_nips19}
X.~Liu, G.~Yin, J.~Shao, X.~Wang, and h.~Li, ``Learning to predict
  layout-to-image conditional convolutions for semantic image synthesis,'' in
  \emph{Advances in Neural Information Processing Systems (NeurIPS)}, 2019.

\bibitem{ulyanov2016instance}
D.~Ulyanov, A.~Vedaldi, and V.~Lempitsky, ``Instance normalization: The missing
  ingredient for fast stylization,'' \emph{arXiv preprint arXiv:1607.08022},
  2016.

\bibitem{kingma2014adam}
D.~P. Kingma and J.~Ba, ``Adam: A method for stochastic optimization,'' in
  \emph{International Conference on Learning Representations (ICLR)}, 2015.

\bibitem{salimans2016improved}
T.~Salimans, I.~Goodfellow, W.~Zaremba, V.~Cheung, A.~Radford, and X.~Chen,
  ``Improved techniques for training gans,'' in \emph{Advances in Neural
  Information Processing Systems (NeurIPS)}, 2016.

\bibitem{nguyen2017dual}
T.~Nguyen, T.~Le, H.~Vu, and D.~Phung, ``Dual discriminator generative
  adversarial nets,'' in \emph{Advances in Neural Information Processing
  Systems (NeurIPS)}, 2017.

\bibitem{Zhou_Wang_tip04}
Z.~Wang, A.~C. Bovik, H.~R. Sheikh, and E.~P. Simoncelli, ``Image quality
  assessment: from error visibility to structural similarity,'' \emph{IEEE
  Transactions on Image Processing (TIP)}, vol.~13, no.~4, pp. 600--612, 2004.

\end{thebibliography}


\begin{IEEEbiography}[{\includegraphics[width=1in,height=1.25in,clip,keepaspectratio]{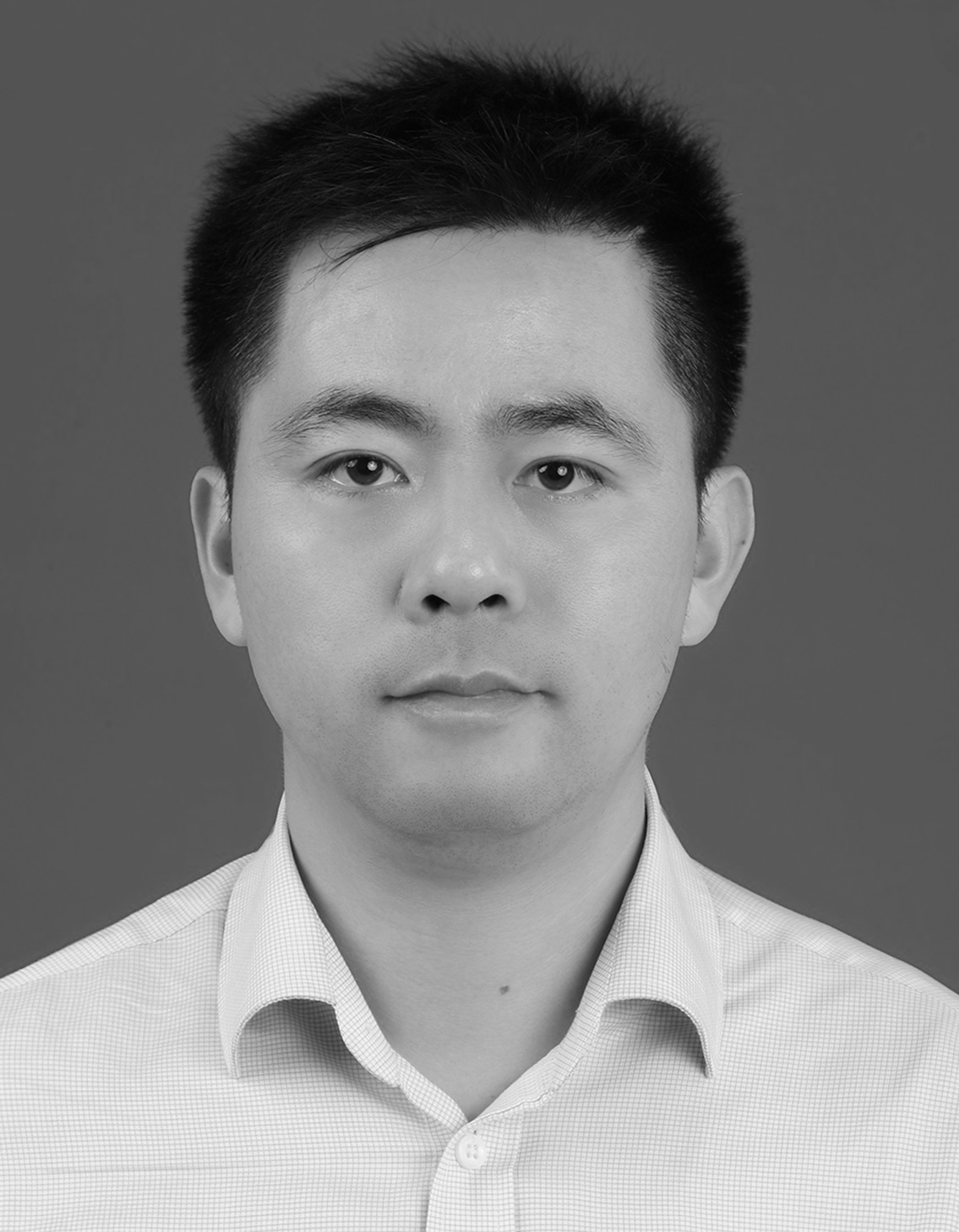}}]{Songsong Wu}
received the Ph.D. degree in pattern recognition and intelligent systems from the Nanjing University of Science and Technology, China, in 2012. He was a Visiting Scientist at Texas State University in San Marco, USA, from 2018 to 2019. He is currently an Assistant Professor at the School of Computer Science, Guangdong University of Petrochemical Technology. His research interests include representation learning, transfer learning, and their applications to pattern recognition, computer vision and fault diagnosis.
\end{IEEEbiography}

\begin{IEEEbiography}[{\includegraphics[width=1in,height=1.25in,clip,keepaspectratio]{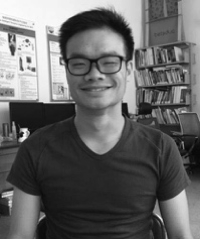}}]{Hao Tang}
is currently a Postdoctoral with Computer Vision Lab, ETH Zurich, Switzerland. He received the master’s degree from the School of Electronics and Computer Engineering, Peking University, China and the Ph.D. degree from Multimedia and Human Understanding Group, University of Trento, Italy. He was a visiting scholar in the Department of Engineering Science at the University of Oxford. His research interests are deep learning, machine learning, and their applications to computer vision.
\end{IEEEbiography}

\begin{IEEEbiography}[{\includegraphics[width=1in,height=1.25in,clip,keepaspectratio]{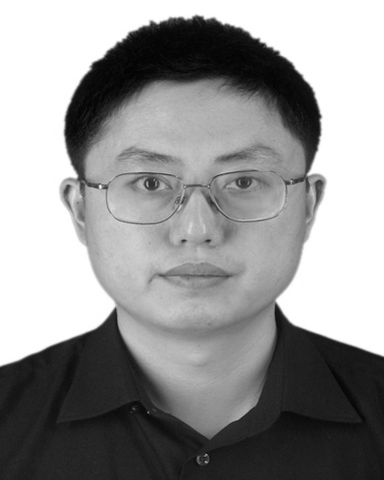}}]{Xiao-Yuan Jing}
received the Ph.D. degree in pattern recognition and intelligent systems from the Nanjing University of Science and Technology, China, in 1998. He was a Professor in the Department of Computer, Shenzhen Research Student School, Harbin Institute of Technology, in 2005. He is currently a Professor and Dean of the School of Computer, Guangdong University of Petrochemical Technology, and a Professor in the School of Computer, Wuhan University, China. He has published more than 100 scientific papers in international journals and conferences such as the IEEE Transactions on Pattern Analysis and Machine Intelligence, IEEE Transactions on Image Processing, IEEE Transactions on Information Forensics and Security, IEEE Transactions on Neural Networks and Learning Systems, TCB, TR, CVPR, AAAI, and IJCAI. His research interests include pattern recognition, machine learning, artificial intelligence, and fault diagnosis.
\end{IEEEbiography}

\begin{IEEEbiography}[{\includegraphics[width=1in,height=1.25in,clip,keepaspectratio]{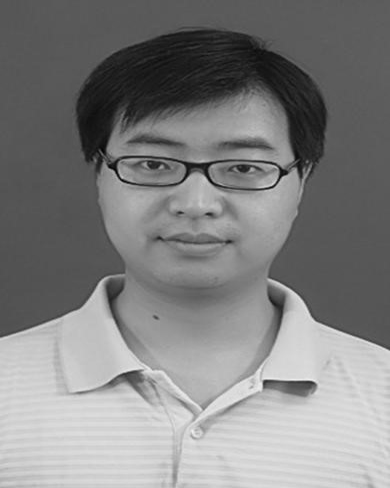}}]{Haifeng Zhao}
received his B.E. and Ph.D. degrees from Nanjing University of Science and Technology, China, in 2005 and 2012 respectively. He is currently a Senior Engineer with School of Software Engineering, Jinling Institute of Technology, China. Before that, he was an Assistant Researcher with  Shenzhen Institutes of Advanced Technology, Chinese Academy of Sciences. He visited the Australian National University and Canberra Research Laboratory of NICTA as a visiting student from 2008 to 2010. He also visited Sydney University and Griffith University in 2017 and 2018. His  research interests include computer vision, pattern recognition and human computer interaction.
\end{IEEEbiography}

\begin{IEEEbiography}[{\includegraphics[width=1in,height=1.25in,clip,keepaspectratio]{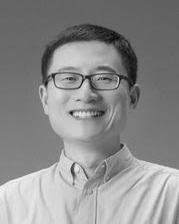}}]{Jianjun Qian}
received the Ph.D. degree in pattern recognition and intelligence systems from the School of Computer Science and Engineering, Nanjing University of Science and Technology (NUST), Nanjing, China, in 2014. He is currently an Associate Professor with the Key Laboratory of Intelligent Perception and Systems for High-Dimensional Information of Ministry of Education, School of Computer Science and Engineering, NUST. His current research interests include pattern recognition, computer vision, and machine learning, especially for face recognition.
\end{IEEEbiography}

\begin{IEEEbiography}[{\includegraphics[width=1in,height=1.25in,clip,keepaspectratio]{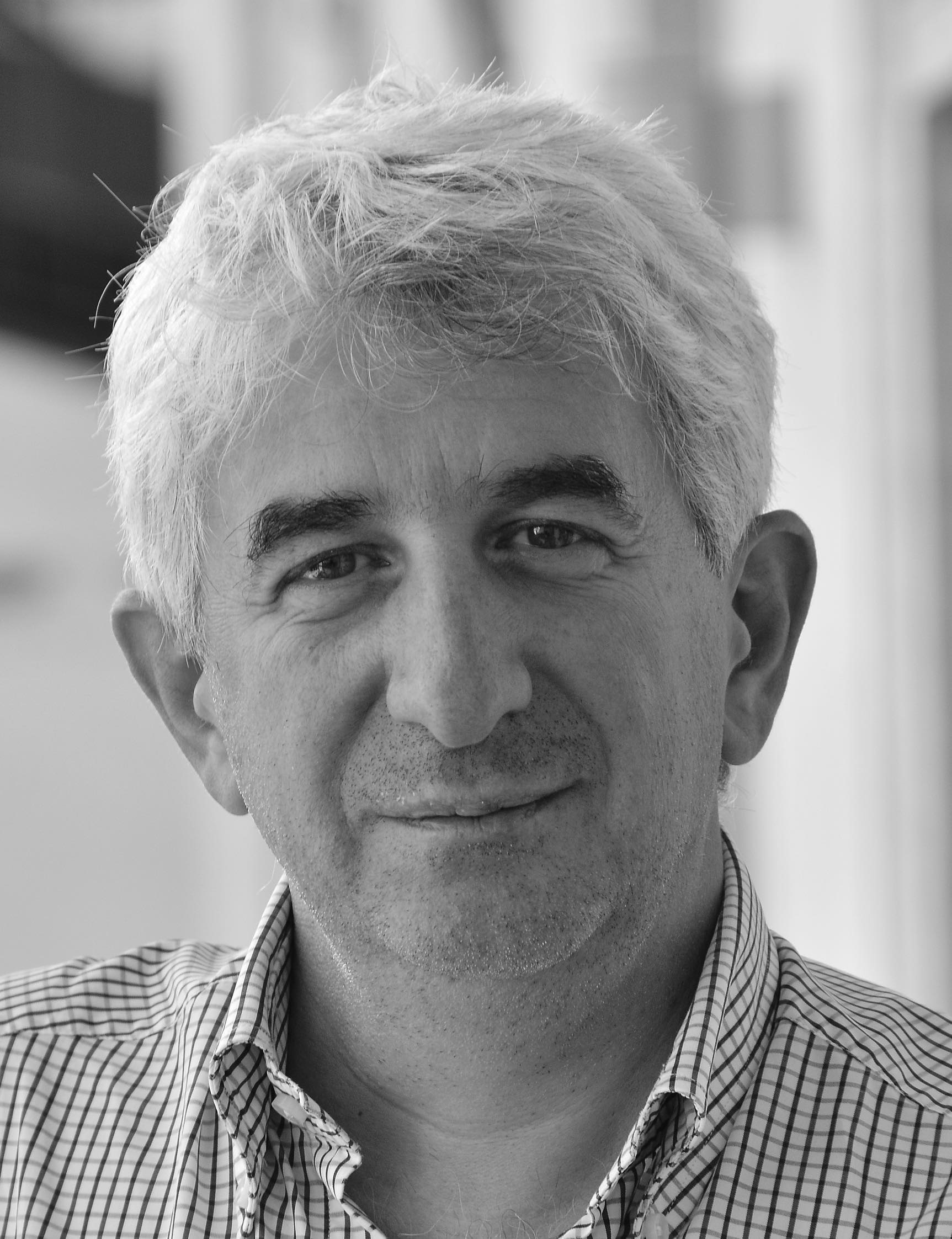}}]{Nicu Sebe}
is Professor with the University of Trento, Italy, leading the research in the areas of multimedia information retrieval and human behavior understanding. He was the General Co-Chair of ACM Multimedia 2013, and the Program Chair of the International Conference on Image and Video Retrieval in 2007 and 2010, ACM Multimedia 2007 and 2011, ICCV 2017 and ECCV 2016. He is a Program Chair of ICPR 2020. He is a fellow of the International Association for Pattern Recognition.
\end{IEEEbiography}

\begin{IEEEbiography}[{\includegraphics[width=1in,height=1.25in,clip,keepaspectratio]{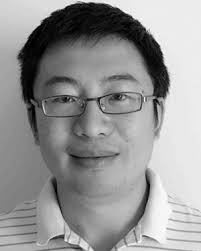}}]{Yan Yan}
received the Ph.D. degree in computer science from the University of Trento. He is currently Gladwin Development Chair Assistant Professor with the Department of Computer Science, Illinois Institute of Technology. He was an Assistant Professor at Texas State University and a Research Fellow at the University of Michigan and the University of Trento. His research interests include computer vision, machine learning, and multimedia.
\end{IEEEbiography}

%\end{IEEEbiography}
%% insert where needed to balance the two columns on the last page with
%%\newpage


%\vfill


\end{document}